\documentclass[lettersize,journal,twoside]{IEEEtran}
\usepackage{amsmath,amsfonts}
\usepackage{algorithmic}
\usepackage{array}
\usepackage[caption=false,font=normalsize,labelfont=sf,textfont=sf]{subfig}
\usepackage{standalone} % 允许 includestandalone
\usepackage{nicematrix}
\usepackage{textcomp}
\usepackage{stfloats}
\usepackage{url}
\usepackage{verbatim}
\usepackage{graphicx}
\usepackage{cite}
\usepackage[table]{xcolor}
\usepackage{listings}
\usepackage{fontawesome5}
\usepackage{hyperref}
\usepackage{multirow}
\usepackage{balance}
\usepackage{CJKutf8}     % 关键：用于支持中文环境
\usepackage{tikz}
\usetikzlibrary{arrows,calc}
\usepackage{adjustbox} % 为了 width=\linewidth 包一层缩放（更稳）

\usepackage{tcolorbox}
\tcbuselibrary{breakable, skins}
\usepackage{enumitem}

\newtcolorbox{highlightbox}[1]{%
    enhanced,%
    colback=cyan!3!white,%
    colframe=cyan!40!black,%
    colbacktitle=cyan!10!white,%
    coltitle=black,%
    fonttitle=\itshape\bfseries,%
    title={#1},%
    halign title=center,%
    arc=3mm, rounded corners,%
    boxrule=0.7pt,%
    left=2mm, right=2mm, top=2mm, bottom=2mm,%
    toptitle=1.5mm, bottomtitle=1.5mm,%
    after skip=10pt,%
    breakable,%
    drop shadow=gray!15%
}

\newenvironment{compactlist}
  {\begin{itemize}[leftmargin=*, noitemsep, topsep=2pt, parsep=0pt, partopsep=0pt]}
  {\end{itemize}}

\usepackage{booktabs}
\usepackage[table]{xcolor}
\usepackage{makecell}
\definecolor{HeaderBlue}{RGB}{214,235,255} % shallow blue
\definecolor{RowGray}{gray}{0.96}

\usepackage{xcolor}
\usepackage{tikz}
\usetikzlibrary{arrows.meta,shapes,positioning,calc,fit,backgrounds}
\usepackage{forest}
\useforestlibrary{edges}

\def\BibTeX{{\rm B\kern-.05em{\sc i\kern-.025em b}\kern-.08em
    T\kern-.1667em\lower.7ex\hbox{E}\kern-.125emX}}

\makeatletter
\renewcommand*\l@section[2]{%
  \@dottedtocline{1}{0em}{2.3em}{\bfseries #1}{#2}
}
\renewcommand*\l@subsection[2]{%
  \@dottedtocline{2}{1.5em}{2.8em}{\small #1}{#2}
}
\renewcommand*\l@subsubsection[2]{%
  \@dottedtocline{3}{3.0em}{3.2em}{\small #1}{#2}
}

\makeatother

\definecolor{uclablue}{rgb}{0.15, 0.45, 0.68}

\definecolor{uclablue}{rgb}{0.15, 0.45, 0.68}

\definecolor{skyblue}{rgb}{0.25, 0.55, 0.85}

\definecolor{mediumblue}{rgb}{0.20, 0.50, 0.78}

\definecolor{cobaltblue}{rgb}{0.24, 0.48, 0.82}

\definecolor{oceanblue}{rgb}{0.18, 0.52, 0.75}

\definecolor{royalblue}{rgb}{0.25, 0.41, 0.88}

\definecolor{dodgerblue}{rgb}{0.12, 0.56, 1.0}

\hypersetup{
    breaklinks,
    colorlinks=true,
    linkcolor=royalblue,    
    citecolor=uclablue,
    urlcolor=uclablue,
}

\begin{document}

\title{
Event Extraction in Large Language Model:\\ A Holistic Survey of Method, Modality, and Future
}
\author{Bobo Li, 
Xudong Han, 
Jiang Liu,
Yuzhe Ding
Liqiang Jing,
Zhaoqi Zhang,
Jinheng Li,\\
Xinya Du,
Fei Li,
Meishan Zhang,
Min Zhang,
Aixin Sun,
Philip S. Yu,
Hao Fei\\

\thanks{Bobo Li, Jinheng Li, and Hao Fei are with the National University of Singapore, Singapore. Xudong Han is with the University of Sussex, U.K. Jiang Liu, Yuzhe Ding, and Fei Li are with Wuhan University, China. Liqiang Jing and Xinya Du are with the University of Texas at Dallas, USA. Zhaoqi Zhang and Aixin Sun are with Nanyang Technological University, Singapore. Meishan Zhang and Min Zhang are with the Harbin Institute of Technology (Shenzhen), China. Philip S. Yu is with the University of Illinois at Chicago, USA. Corresponding author: Hao Fei (e-mail: haofei7419@gmail.com)}

{\faGithub \quad \textbf{\tt \url{https://github.com/unikcc/AwesomeEventExtraction}}}\\
}

\markboth{arXiv preprint}
{Li \MakeLowercase{\textit{et al.}}: Event Extraction}

\maketitle

\vspace{-5mm}
\begin{abstract}

Large language models (LLMs) and multimodal LLMs are changing event extraction (EE): prompting and generation can often produce structured outputs in zero shot or few shot settings. Yet LLM based pipelines face deployment gaps, including hallucinations under weak constraints, fragile temporal and causal linking over long contexts and across documents, and limited long horizon knowledge management within a bounded context window. We argue that EE should be viewed as a system component that provides a cognitive scaffold for LLM centered solutions. Event schemas and slot constraints create interfaces for grounding and verification; event centric structures act as controlled intermediate representations for stepwise reasoning; event links support relation aware retrieval with graph based RAG; and event stores offer updatable episodic and agent memory beyond the context window. This survey covers EE in text and multimodal settings, organizing tasks and taxonomy, tracing method evolution from rule based and neural models to instruction driven and generative frameworks, and summarizing formulations, decoding strategies, architectures, representations, datasets, and evaluation. We also review cross lingual, low resource, and domain specific settings, and highlight open challenges and future directions for reliable event centric systems.
Finally, we outline open challenges and future directions that are central to the LLM era, aiming to evolve EE from static extraction into a structurally reliable, agent ready perception and memory layer for open world systems.

\end{abstract}

\tableofcontents

\newpage

% \usetikzlibrary{calc,fit,decoration,backgrounds,shadows.blur}
\usetikzlibrary{calc,fit,backgrounds,shadows.blur,decorations.pathmorphing}

% 1. 定义新配色
\definecolor{c1}{HTML}{6D9EEB} % 清新蓝
\definecolor{c2}{HTML}{E06666} % 柔和红
\definecolor{c5}{HTML}{F6B26B} % 活力橙
\definecolor{c4}{HTML}{93C47D} % 抹茶绿
% \definecolor{c5}{HTML}{8E7CC3} % 香芋紫
\definecolor{c6}{HTML}{76A5AF} % 青瓷色
\definecolor{mainline}{HTML}{595959} % 主轴深灰

% 初始化默认颜色变量
\newcommand{\evtColor}{c1} 

% 2. 样式设置 (注意：tikzset 内部不要留空行)
\tikzset{
  % --- 样式映射 ---
  sketchy/.style={
    % 1. 填充白色背景 (不画边框，边框交给后面画)
    fill=white, draw=none,
    % 2. 第一遍描边：粗一点
    postaction={
        draw=\evtColor, line width=1pt,
        decorate, decoration={random steps, segment length=3pt, amplitude=0.4pt},
        rounded corners=3pt, line cap=round, line join=round
    },
    % 3. 第二遍描边：细一点，并且稍微偏移一点点 (模拟手抖重描)
    postaction={
        draw=\evtColor, line width=0.6pt,
        decorate, decoration={random steps, segment length=2pt, amplitude=0.3pt},
        rounded corners=2pt, line cap=round, line join=round,
        shift={(0.3pt, 0.1pt)} % 微小的错位
    }},
  handdrawn/.style={
    decorate, 
    decoration={random steps, segment length=5pt, amplitude=0.0pt}
  },
  data/.code={\renewcommand{\evtColor}{c1}},
  ndata/.code={\renewcommand{\evtColor}{c2}},
  % --- 核心组件样式 ---
  evtdot/.style={circle, fill=white, draw=\evtColor, line width=1.5pt, inner sep=1.8pt},
  evtline/.style={draw=\evtColor, line width=1.2pt},
  evtbadge/.style={lab, text=black!80, fill=white, draw=\evtColor, line width=1.2pt, rounded corners=5pt, inner xsep=5pt, inner ysep=3pt,
  % sketchy % <--- 使用双重描边样式
  % handdrawn 
  },
  % --- 字体设置 ---
  lab/.style={align=center, font=\footnotesize\sffamily\bfseries}, 
  year/.style={font=\scriptsize\sffamily},
  % --- Legend 样式 ---
  legend-text/.style={lab, anchor=west, font=\footnotesize\sffamily\bfseries},
  legend-swatch/.style={rounded corners=3pt, line width=1pt, inner sep=0pt, minimum width=15pt, minimum height=10pt},
  legend-group/.style={draw=black!20, dashed, rounded corners=8pt, line width=1pt, inner sep=6pt},
  % --- 兼容性定义 (防止旧代码残留报错) ---
  c1/.style={data}, c2/.style={ndata}, c3/.style={data}, 
  c4/.style={data}, c5/.style={data}, c6/.style={data}
}

\makeatother

% 3. 辅助命令定义
\newcommand{\legendsep}{4pt}
\newcommand{\legsw}{15pt}
\newcommand{\legsh}{12pt}
\newcommand{\legpad}{0pt}

\makeatletter
% 定义 Legend 绘制命令
\newcommand{\legendBetweenNs}[7]{%
  \pt{U}{#2}{#4}%
  \pgfmathsetmacro{\tflip}{1-#4}%
  \pt{L}{#3}{\tflip}%
  \path let \p1=($(U)!0.5!(L)$) in coordinate (#1-pos) at (\p1);
  \node[legend-swatch, draw=#5!90, fill=#5!10, anchor=west] (#1-swatch) at (#1-pos) {};
  \node[legend-text,#7] (#1-text) at ($(#1-swatch.east)+(\legendsep,0)$) {#6};
  \node[inner sep=\legpad, fit=(#1-swatch)(#1-text)] (#1) {};
}

% 定义 Legend 组框
\newcommand{\legendGroup}[4][]{%
  \begin{scope}[on background layer]%
    \node[legend-group,#1, fit=(#3)(#4)] (#2) {};
  \end{scope}%
}

% 定义事件绘制命令 (Up, Down, Left, Right)
% 核心改动：这里使用了 scope 来处理颜色，比老代码更健壮
\newcommand{\eventUp}[5][ndata]{\pt{A}{#2}{#3}%
  \begin{scope}[#1]
    \draw[evtline] (A) node[evtdot]{} -- ++(0,1.4)
      node[evtbadge,above]{#4\\\footnotesize(#5)};
  \end{scope}}

\newcommand{\eventHUp}[5][ndata]{\pt{A}{#2}{#3}%
  \begin{scope}[#1]
    \draw[evtline] (A) node[evtdot]{} -- ++(0,0.4)
      node[evtbadge,above]{#4\\\footnotesize(#5)};
  \end{scope}}

\newcommand{\eventDown}[5][ndata]{\pt{A}{#2}{#3}%
  \begin{scope}[#1]
    \draw[evtline] (A) node[evtdot]{} -- ++(0,-1.4)
      node[evtbadge,below]{#4\\\footnotesize(#5)};
  \end{scope}}

\newcommand{\eventHDown}[5][ndata]{\pt{A}{#2}{#3}%
  \begin{scope}[#1]
    \draw[evtline] (A) node[evtdot]{} -- ++(0,-0.4)
      node[evtbadge,below]{#4\\\footnotesize(#5)};
  \end{scope}}

\newcommand{\eventLeft}[5][ndata]{\pt{A}{#2}{#3}%
  \begin{scope}[#1]
    \draw[evtline] (A) node[evtdot]{} -- ++(-0.4,0)
      node[evtbadge,anchor=east]{#4\\\footnotesize(#5)};
  \end{scope}}

\newcommand{\eventRight}[5][ndata]{\pt{A}{#2}{#3}%
  \begin{scope}[#1]
    \draw[evtline] (A) node[evtdot]{} -- ++(0.4,0)
      node[evtbadge,anchor=west]{#4\\\footnotesize(#5)};
  \end{scope}}
\makeatother

\tikzset{
  my-box/.style={
    rectangle, draw=gray, rounded corners,
    text opacity=1, minimum height=1.5em, minimum width=5em,
    inner sep=2pt, align=center, fill opacity=.5, line width=0.8pt,
  },
  leaf/.style={
    my-box, minimum height=1.5em, fill=pink!10, text=black,
    align=left, font=\normalsize, inner xsep=2pt, inner ysep=4pt,
    line width=0.8pt,
  },
  dot/.style={circle,fill=black,inner sep=1.7pt},
  reddot/.style={circle,fill=red,inner sep=2.1pt},
  % ============== 修改点 1：将 \sffamily 改为 Times New Roman 设置 ==============
  % lab/.style={align=center,font=\footnotesize\sffamily},
  lab/.style={align=center,font=\footnotesize\fontfamily{ptm}\selectfont}, 
  % year/.style={font=\scriptsize\sffamily},
  year/.style={font=\scriptsize\fontfamily{ptm}\selectfont},
  % =========================================================================
}

\begin{figure*}[!t] 
  \centering
  \begin{adjustbox}{width=\linewidth,center}
    \begin{tikzpicture}[
      x=1cm,y=1cm,>=stealth,line cap=round,
      font=\sffamily % 全局使用无衬线字体
    ]

      % 1. 骨架定义 (Skeleton)
      \coordinate (P0) at (0,0);      
      \coordinate (P1) at (20,0);     
      \coordinate (P2) at (20,-6);    
      \coordinate (P3) at (0,-6);     
      \coordinate (P4) at (0,-12);    
      \coordinate (P5) at (20,-12);   

      % 主轴绘制
      \draw[color=mainline, very thick,-stealth] (P0)--(P1)--(P2)--(P3)--(P4)--(P5);

      % 自动计算端点 S1..S5 和 E1..E5
      \foreach \i in {1,2,3,4,5}{
        \coordinate (S\i) at (P\the\numexpr\i-1\relax);
        \coordinate (E\i) at (P\i);
      }

      % 关键工具：\pt 定义 (必须在 tikzpicture 内部)
      \providecommand{\pt}[3]{% name, seg, t
        \path let \p1=($(S#2)!#3!(E#2)$) in coordinate (#1) at (\p1);
      }
      % 这里使用新代码的语法：execute at begin scope
      % 段1：Task Initialization
      \eventDown[execute at begin scope={\def\evtColor{c1}}]    {1}{0.03}{MUC-3, \cite{sundheim91otm}, 1991}{Task Initialization}
      \eventHUp[execute at begin scope={\def\evtColor{c1}}] {1}{0.09}{TimeML\&TimeBank, \cite{pustejovsky03tc}, 2003}{Time Annotation \& Corpus}
      \eventHDown[execute at begin scope={\def\evtColor{c1}}] {1}{0.15}{ACE-2004, \cite{mitchell05a2m}, 2005}{Multilingual Dataset}
      \eventUp[execute at begin scope={\def\evtColor{c1}}] {1}{0.21}{ACE-2005, \cite{walker06a2m}, 2006}{Multilingual Dataset}
      \eventDown[execute at begin scope={\def\evtColor{c2}}]    {1}{0.27}{Stage-Split, \cite{ahn06see}, 2006}{Stage Modeling}
      \eventHUp[execute at begin scope={\def\evtColor{c1}}] {1}{0.33}{BioNLP'09, \cite{kim09obs}, 2009}{BioNLP Dataset}
      \eventHDown[execute at begin scope={\def\evtColor{c2}}] {1}{0.39}{CrossEvent Inference, \cite{liao10dlc}, 2010}{Document Level Inference}
      \eventUp[execute at begin scope={\def\evtColor{c2}}] {1}{0.45}{MaxEnt, \cite{li13jee}, 2013}{Joint Outperform Pipeline}
      \eventDown[execute at begin scope={\def\evtColor{c2}}] {1}{0.51}{DMCNN, \cite{chen15eed}, 2015}{Dynamic CNN}
      \eventHUp[execute at begin scope={\def\evtColor{c2}}] {1}{0.57}{DomainBiLSTM, \cite{nguyen15edd}, 2015}{Domain Adaptation}
      \eventHDown[execute at begin scope={\def\evtColor{c2}}] {1}{0.63}{RichEE, \cite{song15lre}, 2015}{Domain Adaptation}
      \eventUp[execute at begin scope={\def\evtColor{c2}}] {1}{0.69}{JRNN, \cite{nguyen16jee}, 2016}{bidirectional RNN}
      \eventDown[execute at begin scope={\def\evtColor{c2}}] {1}{0.75}{JointEventEntity, \cite{yang16jee}, 2016}{Joint Events \& Entities}
      \eventHUp[execute at begin scope={\def\evtColor{c2}}] {1}{0.81}{FBRNN, \cite{ghaeini16end}, 2016}{BiRNN for Event Detection}
      \eventHDown[execute at begin scope={\def\evtColor{c2}}] {1}{0.87}{GCN-EN, \cite{nguyen18gcn}, 2018}{GCN for Event Detection}
      \eventUp[execute at begin scope={\def\evtColor{c2}}] {1}{0.93}{JMEE, \cite{liu18jme}, 2018}{Attention-based GCN}
      
      % 中间过渡
      \eventLeft[evtbadge/.append style={fill=white,draw=black,text=c2,font=\bfseries}, execute at begin scope={\def\evtColor{c1}}] {2}{0.5}{Bert, \cite{devlin19bpt}, 2018}{PLM Era Begins}

      % 段3：PLM Methods
      \eventUp[execute at begin scope={\def\evtColor{c4}}] {3}{0.08}{HMEAE, \cite{wang19hhm}, 2019}{Hierarchy Modular EE}
      \eventHDown[execute at begin scope={\def\evtColor{c1}}] {3}{0.14}{MAVEN, \cite{wang20m}, 2020}{General Domain Dataset}
      \eventHUp[execute at begin scope={\def\evtColor{c4}}] {3}{0.20}{OneIE, \cite{lin20jnm}, 2020}{Joint EE Framework}
      \eventDown[execute at begin scope={\def\evtColor{c1}}] {3}{0.26}{M2E2, \cite{li20cross}, 2020}{Multimedia EE Dataset}
      \eventUp[execute at begin scope={\def\evtColor{c1}}] {3}{0.32}{RAME, \cite{ebner20msa}, 2020}{Document-level Arg Data}
      \eventHDown[execute at begin scope={\def\evtColor{c4}}] {3}{0.38}{EEMRC, \cite{liu20eem}, 2020}{EE as MRC}
      \eventHUp[execute at begin scope={\def\evtColor{c4}}] {3}{0.44}{BERT-QA, \cite{du20eea}, 2020}{QA-style EE}
      \eventDown[execute at begin scope={\def\evtColor{c4}}] {3}{0.50}{CASIE, \cite{satyapanich20cec}, 2020}{Cybersecurity EE Data} 
      \eventUp[execute at begin scope={\def\evtColor{c4}}] {3}{0.56}{MGR, \cite{du20dle}, 2020}{Role Filler EE} 
      \eventHDown[execute at begin scope={\def\evtColor{c1}}] {3}{0.62}{WikiEvents, \cite{li21dle}, 2021}{Document Event} 
      \eventHUp[execute at begin scope={\def\evtColor{c1}}] {3}{0.68}{VM2E2, \cite{chen21jme}, 2021}{Video EE Data} 
      \eventDown[execute at begin scope={\def\evtColor{c4}}] {3}{0.74}{Text2Event, \cite{lu21tcs}, 2021}{Structure Generation} 
      \eventUp[execute at begin scope={\def\evtColor{c4}}] {3}{0.80}{TANL, \cite{paolini21spt}, 2021}{EE as Translation} 
      \eventHDown[execute at begin scope={\def\evtColor{c4}}] {3}{0.86}{DEGREE, \cite{hsu22dde}, 2022}{Generative-based EE} 
      \eventHUp[execute at begin scope={\def\evtColor{c4}}] {3}{0.92}{OneEE, \cite{cao22oos}, 2022}{Grid-Tagging EE} 
      
      % 中间过渡
      \eventRight[evtbadge/.append style={fill=white,draw=black,text=c2,font=\bfseries}, execute at begin scope={\def\evtColor{c1}}] {4}{0.5}{ChatGPT, \cite{brown20lma}, 2022}{LLM Era Begins}

      % 段5：LLM Methods
      \eventHDown[execute at begin scope={\def\evtColor{c5}}] {5}{0.03}{ChatIE, \cite{wei23czs}, 2023}{ChatGPT for EE} 
      \eventUp[execute at begin scope={\def\evtColor{c5}}] {5}{0.09}{OmniEvent, \cite{peng23ocf}, 2023}{LLM-based EE Tools} 
      \eventDown[execute at begin scope={\def\evtColor{c5}}] {5}{0.15}{CAMEL, \cite{du23tme}, 2023}{Generation Argument EE} 
      \eventHUp[execute at begin scope={\def\evtColor{c5}}] {5}{0.21}{MosaiCLIP, \cite{singh23cfc}, 2023}{Contrastive Learning for MMEE} 
      \eventHDown[execute at begin scope={\def\evtColor{c5}}] {5}{0.27}{InstructUIE, \cite{wang23imt}, 2023}{Instruction-tuned EE}
      \eventUp[execute at begin scope={\def\evtColor{c1}}] {5}{0.33}{DIE-EC, \cite{gao24ece}, 2024}{Cross-Document ECR Data} 
      \eventDown[execute at begin scope={\def\evtColor{c1}}] {5}{0.39}{CDEE, \cite{gao24hem}, 2024}{Cross-Document EE Data} 
      \eventHUp[execute at begin scope={\def\evtColor{c1}}] {5}{0.45}{DiscourseEE, \cite{sharif24eis}, 2024}{Implicit EE Data} 
      \eventHDown[execute at begin scope={\def\evtColor{c5}}] {5}{0.51}{HD-LoA, \cite{zhou24llt}, 2024}{In-Context Learning for EE} 
      \eventUp[execute at begin scope={\def\evtColor{c5}}] {5}{0.57}{ULTRA, \cite{zhang24uul}, 2024}{Enhance LLM for EE} 
      \eventDown[execute at begin scope={\def\evtColor{c5}}] {5}{0.63}{UMIE, \cite{sun24uum}, 2024}{Instruction-tuned MMEE} 
      \eventHUp[execute at begin scope={\def\evtColor{c5}}] {5}{0.69}{GEMS, \cite{lin25gge}, 2025}{Ontology-guided EE} 
      \eventHDown[execute at begin scope={\def\evtColor{c5}}] {5}{0.75}{XMTL, \cite{cao25cmm}, 2025}{Multi-task Learning for EE} 
      \eventUp[execute at begin scope={\def\evtColor{c5}}] {5}{0.81}{LSED, \cite{qiu25tia}, 2025}{LLM for Social EE} 
      \eventDown[execute at begin scope={\def\evtColor{c5}}] {5}{0.87}{Sed-Aug, \cite{ma25eid}, 2025}{Data Augmentation for EE} 
      \eventHUp[execute at begin scope={\def\evtColor{c5}}] {5}{0.93}{SEOE, \cite{lu25s}, 2025}{Open Domain EE} 

      % =========================================================
      % 3. 绘制图例 (Legends)
      % =========================================================
      \legendBetweenNs{leg13A}{1}{3}{0.27}{c1}{Dataset}
        {font=\large\bfseries, align=left, inner sep=0pt, outer sep=0pt}
      \legendBetweenNs{leg13B}{1}{3}{0.62}{c2}{Pre-LLM Methods}
        {font=\large\bfseries, align=left, inner sep=0pt, outer sep=0pt}
      \legendGroup[draw=black!70,  line width=1.9pt,densely dashed]{g13}{leg13A}{leg13B}

      \legendBetweenNs{leg35A}{3}{5}{0.40}{c4}{PLM-based Methods}
        {font=\large\bfseries}
      \legendBetweenNs{leg35B}{3}{5}{0.73}{c5}{LLM-based Methods}
        {font=\large\bfseries, text width=15em, align=left}
      \legendGroup[draw=black!70, line width=1.9pt, densely dashed]{g35}{leg35A}{leg35B}

    \end{tikzpicture}
  \end{adjustbox}
  \caption{A roadmap of Event Extraction (1991–2025): Key milestones in the Pre-PLM, PLM, and LLM eras.}
  \label{fig:timeline}
\end{figure*}
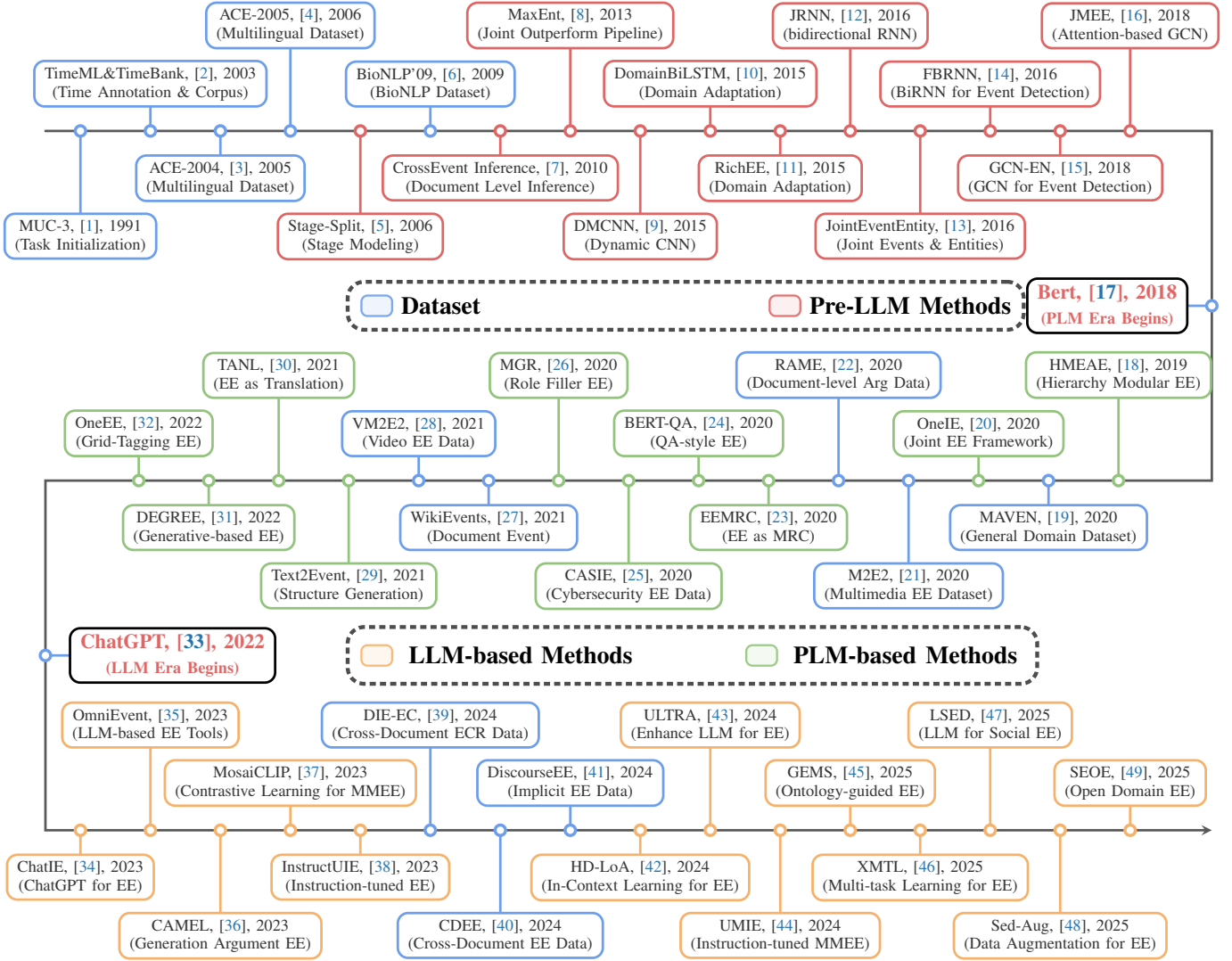

\section{Introduction}
\IEEEPARstart{E}vent extraction (EE) is a core task in natural language processing that aims to identify event triggers, event types, and participant roles from unstructured text, and to organize them into a computable structured representation \cite{li21dle}. Unlike static facts at the entity or relation level, events capture what happened, who was involved, when and where it happened, how it unfolded, and what outcomes followed. This capability is important in application settings that require tracking and interpreting real-world dynamics, including financial risk control and public opinion monitoring, clinical course tracking, situational awareness, and public safety and emergency warning. Over the past two decades, the community has developed many datasets and benchmarks, and has advanced methods from rule-based and feature-engineering approaches to neural and graph-based modeling \cite{nguyen16jee,liu18jme,li21dle}. These efforts have also supported the construction and use of event knowledge bases and event graphs, making EE a key pillar within the broader information extraction landscape.

The rise of large language models (LLMs) is reshaping the practice of information extraction. Models that previously required task-specific training can now often be replaced by prompting a general-purpose LLM to directly produce outputs that resemble structured records, sometimes even in zero-shot or few-shot settings \cite{openai23gtr,deepseek25dr}. This shift raises an unavoidable question: \textbf{in an era where LLMs can process text end to end and generate structured outputs, is event extraction still necessary as an independent research direction?} In many real deployments, the prevailing practice is increasingly to feed raw text to an LLM, rather than to first extract structured events and then perform downstream reasoning and decision making on top of those structures. As a result, EE may appear less central than before, and it can be mistakenly viewed as something that end-to-end generation can simply replace.

This survey argues that LLMs do not diminish the value of event extraction. Instead, they push EE from a task- or model-centric problem toward a system-level structured interface and constraint layer. The key distinction is that practical deployments care not only about producing an answer, but also about meeting system requirements such as reliability, traceability, and long-horizon knowledge management. Under these requirements, relying solely on unconstrained end-to-end generation exposes substantial \emph{cognitive gaps}. First, generative outputs are probabilistic. Without explicit structural constraints, models can hallucinate and errors can accumulate across multi-step pipelines \cite{huang23hall}. Second, when evidence is dispersed across long contexts or multiple documents, models often struggle to maintain stable links among temporal order, causal chains, and role coreference. This makes the reasoning process brittle and difficult to audit. Third, similarity-based retrieval does not guarantee access to precise temporal or causal relations, and limited context windows cannot accommodate a continuous stream of experiences in open environments. Consequently, simply stacking more text is often insufficient for long-term planning and consistent behavior.

Event extraction offers a structured complement that directly targets these system-level gaps. Because EE outputs are explicit, constrained, and computable, they can serve as intermediate representations and external memories in LLM-centered systems. In this sense, EE evolves from a static prediction task into a \emph{cognitive scaffold}. First, for reliability, event schemas and slot constraints provide concrete interfaces for grounding and verification, narrowing the space of free-form generation and supplying anchors for checking and correction. Second, for reasoning, event chains decompose narratives into discrete steps and can function as controlled intermediate structures analogous to Chain-of-Thought reasoning, improving controllability and reproducibility \cite{wei22cot}. Third, for knowledge access and memory, events and their temporal, causal, and role links enable retrieval to move beyond similarity matching toward relation-navigable, graph-based retrieval-augmented generation \cite{lewis20rag,peng24gragsurvey}. This organization further supports updatable episodic memory, which is useful for agents that require long-horizon planning without being constrained by context overflow \cite{park23ga}. Therefore, in the LLM era, the value of EE lies less in being the only path to structured outputs, and more in providing a structural backbone for verification, reasoning, retrieval, and agent memory.

Motivated by this perspective, this paper revisits event extraction along the axis of \emph{tasks, datasets and evaluation, and methodological paradigms}. We start from the classic definitions and decompositions of textual EE, review representative datasets, evaluation protocols, and metrics, and then summarize the evolution of modeling approaches from rule-based and traditional learning methods to neural, generative, and instruction-driven frameworks. We further discuss how multimodal and cross-document settings extend the boundary of EE. Building on this foundation, we examine how emerging generative models and agentic systems reshape the functional role of EE in practice, highlighting common collaboration patterns with LLMs, including structured constraints, verifiable workflows, graph-based retrieval, and external memory. Finally, we distill open challenges and future directions guided by application requirements, aiming to inform the design of reliable, controllable, and deployable event-centric intelligent systems.

The remainder of this survey is organized as follows. Section~II introduces the task definition and taxonomy of event extraction, unifies the core subtasks in textual EE, and extends the discussion to multimodal settings such as vision, video, and speech. Section~III to Section~VI review methods and modeling routes, covering rule-based and traditional learning approaches, deep learning, and the instruction-driven and generative paradigms in the LLM and multimodal LLM era, together with common task formulations, decoding strategies, system architectures, and representation design. Section~VII summarizes datasets, evaluation metrics, and toolchains, and Section~VIII discusses diverse application settings such as cross-lingual and low-resource scenarios, different granularities, and vertical domains. Section~IX outlines open challenges and future directions, and Section~X concludes the survey.

\begin{highlightbox}{Evolution: Static Task vs. Cognitive Scaffold}
\begin{compactlist}
    \item Traditionally, EE was viewed as a standalone prediction task to populate static knowledge bases.
    \item In the LLM era, EE evolves into a structural interface (Scaffold) that enhances system reliability, reasoning, and memory.
\end{compactlist}
\end{highlightbox}

\section{Tasks \& Taxonomy}
\label{sec:task}

\begin{figure*}[!t]
% ================= COLOR PALETTE =================
\definecolor{c1}{HTML}{7A7A73} % Gray
\definecolor{c2}{RGB}{237,110,106} % Red
\definecolor{c3}{RGB}{240,154,69}  % Orange
\definecolor{c4}{RGB}{8,153,68}    % Green

\centering
\resizebox{\textwidth}{!}{
\begin{forest}
% ================= TREE CONFIGURATION =================
forked edges,
for tree={
    grow=east,
    reversed=true,
    anchor=west,
    parent anchor=east,
    child anchor=west,
    base=center,
    font=\sffamily\large,
    rectangle,
    draw=gray,
    rounded corners=2pt,
    minimum width=6em,
    edge+={darkgray, line width=1pt},
    s sep=10pt,
    inner xsep=6pt,
    inner ysep=6pt,
    line width=0.8pt,
},
% ================= NODE STYLES =================
ver/.style={
    rotate=90,
    child anchor=north,
    parent anchor=south,
    anchor=center,
    font=\sffamily\Large\bfseries,
    inner sep=6pt
},
leafnode/.style={
    fill=white,
    draw=#1,
    line width=0.8mm,
    edge={#1, line width=1pt},
    align=left,
    text width=58em,        % <--- 关键：加宽宽度，确保一行能放8个
    font=\sffamily\normalsize
},
structnode/.style={
    fill=#1!15,
    draw=#1,
    line width=0.8pt,
    edge={#1, line width=1pt},
    font=\sffamily\bfseries
},
% ================= LEVEL SETTINGS =================
where level=1{text width=10em, text centered}{},
where level=2{text width=11em, text centered}{},
where level=3{text width=60em}{}, 
%
% ================= TREE CONTENT =================
[
    \shortstack{\textbf{Event Extraction}\\\textbf{in LLM Era}}, ver, line width=1mm, draw=black!70
    %
    % ========= BRANCH 1: TASK DIMENSIONS =========
    [
    \textbf{Task Dimensions}, structnode=c1
        [
        \textbf{Extraction Scope}, structnode=c1
            [ {Sentence: ACE05 \cite{doddington04ace1}, FewEvent \cite{deng20mld}, PHEE \cite{sun22p}, MAVEN \cite{wang20m}, CASIE \cite{satyapanich20cec}, GENEVA \cite{parekh23g}, RAMS \cite{ebner20msa}, MINION \cite{pouranbenveyseh22m1} \\
            Document: Doc2EDAG \cite{zheng19dee}, WIKIEVENTS \cite{li21dle}, DE-PPN \cite{yang21dle}, DocEE \cite{wang22d}, RoleEE \cite{jiao22ova}, FEED \cite{li22fcf}, DuEE \cite{li20dls}, HBTNGMA \cite{chen18ced} \\
            Cross-Doc: EventStoryLine \cite{caselli16sar}, WEC-Eng \cite{eirew21w}, RECB \cite{zhao25bbb}, ECB+ \cite{cybulska14scn}, MCECR \cite{pouranbenveyseh24m}, FCC \cite{bugert21gcd}, LegalCore \cite{wei25l}, GVA \cite{vossen18d}}, leafnode=c1 ]
        ]
        [
        \textbf{Event Semantics}, structnode=c1
            [ {Coreference: MAVEN-ERE \cite{wang20m}, KBP2017 \cite{choubey17tak}, GraphECR \cite{chen09gec}, WEC-Zh \cite{gao24ece}, Joint ECR \cite{lee12jee}, RJoint ECR \cite{barhom19rjm}, DIE-EC \cite{gao24hem}  \\
            Relations: MATRES \cite{ning18maa}, Causal-TimeBank \cite{mirza14acb}, HiEve \cite{glavavs14h}, TimeBank-Dense \cite{cassidy14afd}, TDDiscourse \cite{naik19t}, DocRED \cite{wang22d}, MECI \cite{lai22m} \\
            Ontology: RAMS \cite{ebner20msa}, PHEE \cite{sun22p}, OntoEvent \cite{deng21o}, RichERE \cite{song15lre1}, MACCROBAT-EE \cite{ma23d}, MLEE \cite{pyysalo12eea}, MUC-4 \cite{sundheim92of}}, leafnode=c1 ]
        ]
        [
        \textbf{Multimodality}, structnode=c1
            [ {Visual/Video: imSitu \cite{yatskar16srv}, VidSitu \cite{sadhu21vsr}, SWiG \cite{pratt20gsr}, VASR \cite{bitton23vva}, Grounded VidSitu \cite{khan22gvs}, Video Swin \cite{liu22vst}, SlowFast \cite{feichtenhofer19snv}, I3D \cite{carreira17qva} \\
            Audio/Speech: SpeechEE \cite{wang24snb}, DeepSpeech \cite{ghannay18een}, VGGish \cite{chiba21dsr}, wav2vec 2.0 \cite{baevski20w2f}, Whisper \cite{radford23rsr}, ASR-based \cite{wu22res} \\
            Cross-modal: M\(^2\)E\(^2\) \cite{li20cross}, VOANews \cite{shih22cec}, MUIE \cite{zhang24rea}, CMMEvent \cite{liu25cme}, VM\(^2\)E\(^2\) \cite{chen21jme}, MultiHiEve \cite{ayyubi24bge}, M-VAE \cite{ma25sms}}, leafnode=c1 ]
        ]
    ]
    %
    % ========= BRANCH 2: MODELING PARADIGMS =========
    [
    \textbf{Modeling Paradigms}, structnode=c2
        [
        \textbf{Discriminative}, structnode=c2
            [ {Sequence Labeling: OneIE \cite{lin20jnm}, DCFEE \cite{yang18d}, BEESL \cite{ramponi20bee}, ILP \cite{zeng18sue}, SP-IE \cite{lu12aee}, SSED \cite{ferguson18sse}, MTTLADE \cite{elallaly21mmt}, DLRNN \cite{duan17edl} \\
            Span/Pointer: DyGIE++ \cite{wadden19ere}, PLMEE \cite{yang19ept}, MQAEE \cite{li20eem}, BERT-QA \cite{du20eea}, PAIE \cite{ma22pep}, RCEE\_ER \cite{liu20eem}, CasEE \cite{sheng21cjl}, RAAT \cite{liang22rra} \\
            Graph/GNN: JMEE \cite{liu18jme}, S-CNNs \cite{zhang16jee}, JRNN \cite{nguyen16jee}, DMCNN \cite{chen15eed}, PMCNN \cite{li18ebe}, DBRNN \cite{sha18jee}, JointEventEntity \cite{yang16jee}}, leafnode=c2 ]
        ]
        [
        \textbf{Generative}, structnode=c2
            [ {Seq2Seq: Text2Event \cite{lu21tcs}, UIE \cite{lu22usg}, LasUIE \cite{fei22lasuie}, OneEE \cite{cao22oos}, EDEE \cite{wan23jdl}, DualCor \cite{cui22ece}, ODEE \cite{ning23oos}, JEF-HM \cite{pu24jfh} \\
            Prompting: BART-Gen \cite{li21dle}, DEGREE \cite{hsu22dde}, PAIE \cite{ma22pep}, GTEE-DynPref \cite{liu22dpt}, DE-PPN \cite{yang21dle}, SCPRG \cite{liu23edl}, AMPERE \cite{hsu23aaa} \\
            Unified Models: DeepStruct \cite{wang22d}, InstructUIE \cite{wang23imt}, OmniEvent \cite{peng23ocf}, ADELIE \cite{qi24a}, TimeLlaMA \cite{yuan24back}, RLQG \cite{hong24bqg1}, CollabKG \cite{wei24clh}}, leafnode=c2 ]
        ]
        [
        \textbf{LLM-Specific}, structnode=c2
            [ {Instruction Tuning: InstructUIE \cite{wang23imt}, ChatUIE \cite{wang23c}, LLMERE \cite{hu25llm}, ITAG \cite{srivastava25itl}, ECIE \cite{li23eci}, LLMR \cite{ma23llm}, GLF \cite{pang23glc}, GCIE \cite{bao24gcf} \\
            Code Generation: Code4Struct \cite{wang23c}, CodeIE \cite{wang23c}, GoLLIE \cite{sainz24gl}, LC4EE \cite{zhu24l}, RUIE \cite{liao25r}\\
            Agent/Reasoning: ChatIE \cite{wei23czs}, STAR \cite{ma24sbl},  LLM-ERL \cite{chen24ill}, DAO \cite{wang24doa}, TALOR-EE \cite{wang24tal}, SBDA \cite{jin24sda}}, leafnode=c2 ]
        ]
    ]
    %
    % ========= BRANCH 3: LEARNING STRATEGIES =========
    [
    \textbf{Learning Strategies}, structnode=c3
        [
        \textbf{Data Efficiency}, structnode=c3
            [ {Few-Shot: FewEvent \cite{deng20mld}, FewFC \cite{zhou21wrv}, FewDocAE \cite{yang22fsd}, ZSEE \cite{he24z}, Title2Event \cite{deng22t} \\
            Unsupervised: CrossCluster \cite{ji08ree}, Type-Induction \cite{huang23tbr}, R2E \cite{dutkiewicz14rre}, Genia 2011 \cite{kim11o}}, leafnode=c3 ]
        ]
        [
        \textbf{Knowledge Augmented}, structnode=c3
            [ {Retrieval (RAG): RCEE\_ER \cite{liu20eem}, GREE \cite{huang23eed}, GRIT \cite{du20ggr}, GENEVA \cite{parekh22gbg}, TOPJUDGE \cite{feng22ljp}, MoRAG-FD \cite{zhao242ii}, EAESR \cite{shuang24tah} \\
            Schema/Gloss: GEANet \cite{huang20bee}, SROLEPRED \cite{jiao22ova}, SemSynGTN \cite{veyseh20gtn}, BRAD \cite{lai21eeh}, Seq2EG \cite{sun23sne}}, leafnode=c3 ]
        ]
        [
        \textbf{Representation}, structnode=c3
            [ {Embedding: BERT \cite{devlin19bpt}, SpanBERT \cite{joshi20sip}, RoBERTa \cite{liu19rro}, BART \cite{lewis19bds}, GATE \cite{ahmad21gga}, EABERT \cite{xi23eea}, Mabert \cite{ding23mma}, PLMEE \cite{yang19ept} \\
            Syntax/Structure: TSAR \cite{xu22tsa}, AMPERE \cite{hsu23aaa}, TAG \cite{yang23alp}, EPIG-EAE \cite{wan25epi}, GTEE-DynPref \cite{liu22dpt}, PAIE \cite{ma22pep}, SemSynGTN \cite{veyseh20gtn}}, leafnode=c3 ]
        ]
    ]
    %
    % ========= BRANCH 4: ECO-SYSTEM =========
    [
    \textbf{Eco-system}, structnode=c4
        [
        \textbf{Domain Applications}, structnode=c4
            [ {Biomedical: PHEE \cite{sun22p}, DeepEventMine \cite{trieu20dee}, SPEED++ \cite{parekh24s}, BioNLP \cite{kim11o}, BioDEX \cite{d23b}, MACCROBAT-EE \cite{ma23d}, AniEE \cite{kim23a} \\
            Finance: FinEvent \cite{peng22ric}, Doc2EDAG \cite{wang22d}, CFinDEE \cite{zhang24ccf}, DCFEE \cite{yang18d}, ChFinAnn \cite{wang22d}, OEE-CFC \cite{wan24o}, CFinEE \cite{wu23cfg} \\
            Legal/Cyber: LEVEN \cite{yao22lls}, CASIE \cite{satyapanich20cec}, LEEC \cite{xue23lle}, LegalCore \cite{wei25l}, ExcavatorCovid \cite{min21eee}, AEs \cite{guellil25aee}, SEOE \cite{lu25s}}, leafnode=c4 ]
        ]
        [
        \textbf{Tools \& Platforms}, structnode=c4
            [ {Frameworks: TextEE \cite{huang23tbr}, OneEE \cite{cao22oos}, DeepStruct \cite{wang22d}, OmniEvent \cite{peng23ocf}, DeepKE \cite{zhang22ddl}, LFDe \cite{kan24llf} \\
            Integrated Models: DyGIE++ \cite{wadden19ere}, SpERT \cite{eberts19sje}, RESIN \cite{wen21rds}, CasEE \cite{sheng21cjl}, UIE \cite{lu22usg}, CollabKG \cite{wei24clh}, ChatUIE \cite{wang23c}}, leafnode=c4 ]
        ]
        [
        \textbf{Evaluation}, structnode=c4
            [ {Standard: Strict/Soft Match F1, Trigger/Argument ID, Head/Exact Match (ACE05/MAVEN metrics), Mention Detection Rate \\
            Advanced: Coreference (MUC/B$^3$/CEAF/BLANC), Temporal F1, Hallucination Rate, Cross-modal Alignment Score}, leafnode=c4 ]
        ]
    ]
]
\end{forest}
}
\caption{Taxonomy of Event Extraction research in the LLM era, structured by tasks, modeling paradigms, learning strategies, and ecosystem.}
\label{fig:ee-taxonomy}
\end{figure*}
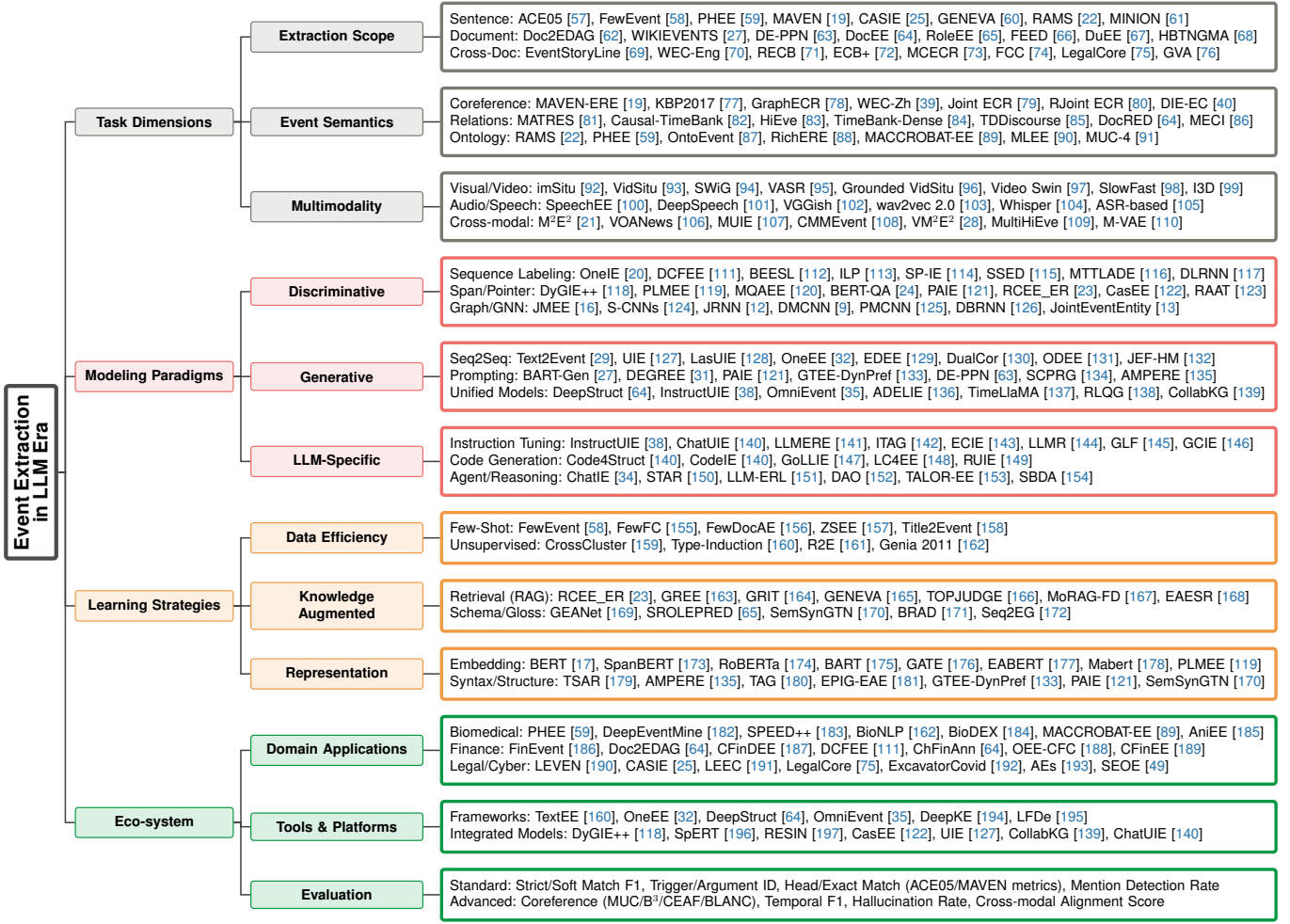

To systematically navigate the diverse landscape of Event Extraction (EE), we present a comprehensive taxonomy in Fig.~\ref{fig:ee-taxonomy}, which organizes the field into tasks, methods, paradigms, and applications.
In this section, we focus specifically on the \textbf{Tasks} dimension to define the fundamental problem scope.
We categorize EE tasks into two primary streams: the established \textit{Text-based EE}, comprising subtasks like trigger detection and argument extraction, and the expanding \textit{Multimodal EE}, which incorporates visual and acoustic signals.

\begin{figure*}[t]
\centering

\subfloat[Event Trigger Detection\label{fig:text:trigger}]{
  \includegraphics[width=0.45\textwidth]{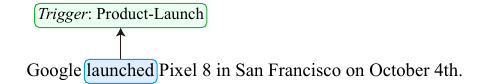}
}\hfill
\subfloat[Event Extraction\label{fig:text:extraction}]{
  \includegraphics[width=0.45\textwidth]{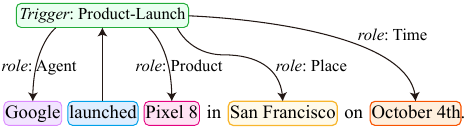}
}

\par\smallskip

\subfloat[Event Relation Extraction\label{fig:text:relation}]{
  \includegraphics[width=0.45\textwidth]{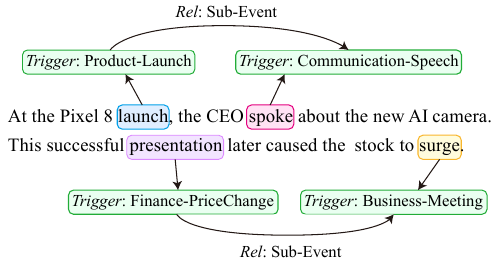}
}\hfill
\subfloat[Event Coreference Resolution\label{fig:text:coreference}]{
  \includegraphics[width=0.45\textwidth]{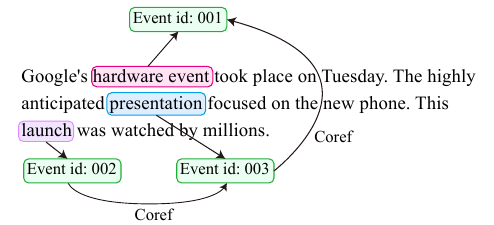}
}

\caption{Task illustration for Text-based Event Extraction.}
\label{fig:task}
\end{figure*}

\subsection{Text-based EE}
Text-based event extraction~\cite{hogenboom11oee, sainz24gl, nguyen16jee} is the most established and widely studied setting in the broader event understanding landscape.
The objective is to transform unstructured text into structured event knowledge—typically, machine-readable records that specify an event type, its participants, attributes, and links to other events.
In practice, the task is commonly decomposed into a pipeline of interconnected sub-tasks: trigger detection~\cite{le21fge}, argument extraction \cite{wang19hhm}, event coreference resolution~\cite{lee12jee}, and event-to-event relation extraction~\cite{wang20jcl}.
Each stage feeds forward signals that refine subsequent stages, while later stages provide opportunities for joint or global consistency checks over earlier decisions.

\subsubsection{Trigger Detection \& Typing}

Trigger detection and typing~\cite{chen18ced, le21fge} is the foundational step in event extraction: the system identifies the text spans that most clearly signal an event’s occurrence and assigns them a type within an event schema.
Triggers are the lexical anchors of an event mention.
As shown in Fig.~\ref{fig:text:trigger}, they are often verbs (e.g., “attacked,” “resigned”) or nominalizations (e.g., “an attack,” “the resignation”), but can also include adjectives or multiword expressions depending on the schema and domain.

Successfully identifying and typing these triggers, however, involves overcoming several core challenges. First, the nature of the event schema itself is a primary consideration. Closed schemas with a predefined ontology (e.g., the types in ACE-style resources~\cite{walker06a2m}) constrain the label space and facilitate supervised learning, whereas open schemas demand models that can generalize to previously unseen or fluid event types discovered from data. Second, further complexity arises from the linguistic form of triggers. Beyond single tokens, they can manifest as multiword expressions (e.g., “blew up”) or even discontinuous phrases (e.g., “set … off”), which challenges token-contiguous span assumptions and motivates more sophisticated structured prediction. Finally, a significant challenge lies in handling nested events, such as “the announcement of a merger.” These hierarchical structures strain flat, span-based tagging approaches, requiring systems that can recognize such compositions and maintain consistent typing across levels~\cite{cao22oos}.

\subsubsection{Argument Extraction}

Argument extraction~\cite{wang19hhm} identifies the participants and attributes associated with a detected event and assigns them semantically meaningful roles.
As shown in Fig.~\ref{fig:text:extraction}, given a specific trigger, the system selects text spans, typically entity mentions, temporal expressions, and locations, and links them as arguments to that event, thereby answering the ``who, what, when, and where'' of the event mention~\cite{liu17eai, li21dle, wang19hhm}.

Accurately extracting these arguments poses several significant challenges.
A primary task is to assign the correct semantic roles (e.g., \textit{Attacker}, \textit{Victim}, \textit{Place}) to each argument based on a predefined schema.
This process requires sophisticated role disambiguation, which becomes particularly difficult when dealing with label imbalance or domain shifts.
The complexity is further compounded because arguments do not always appear adjacent to the event trigger.
Participants may be referenced elsewhere in the text through anaphora (e.g., pronouns) or bridging descriptions, compelling robust systems to leverage coreference resolution and broader discourse context.
Finally, ensuring the semantic integrity of the extracted structure is crucial.
A single entity may legitimately fill multiple roles, and schemas often impose strict type constraints (e.g., a \textit{Victim} must be a person).
Addressing these overlaps and enforcing constraints often requires models capable of joint inference and applying global logic across the document.

\subsubsection{Event Coreference Resolution}

Event coreference resolution~\cite{lee12jee, chen09gec, liu18gbd} clusters event mentions that denote the same underlying occurrence.
As shown in Fig.~\ref{fig:text:coreference}, the goal is to aggregate information, track event evolution across mentions, and avoid redundant or fragmented representations by merging co-referential mentions into a single canonical event~\cite{barhom19rjm, lu18ecr}.

This task is fundamentally challenging due to the multi-dimensional nature of coreference signals and the varying scope of analysis. A central difficulty is that signals for coreference span across triggers, arguments, and attributes. For triggers, this involves matching lexical identity, morphological variants, and semantic similarities. For arguments, it requires assessing the compatibility of key roles, such as shared participants or locations. For attributes, it demands consistency in time, polarity, and modality. Reconciling partial overlaps—where mentions share some but not all of these features—remains a key hurdle. The complexity is further shaped by the task's scope. Within a single document (intra-document), resolution can leverage strong local discourse cues, while across multiple documents (cross-document), systems must contend with greater lexical variability and temporal drift, necessitating more robust normalization and external knowledge.

\subsubsection{Event--Event Relations}
As presented in Fig.~\ref{fig:text:relation}, event-to-event relation extraction~\cite{wang20jcl, chen02em, wadden19ere} identifies structural links among events, enabling timeline construction, causal interpretation, and narrative composition.
This is a higher-level reasoning task that abstracts beyond individual mentions and frames to infer relations that organize events into interpretable structures.

These relations are commonly categorized into three primary types. First, temporal relations~\cite{wang17ioi} order events along a timeline. This entails temporal anchoring—linking events to absolute time expressions and normalizing them (e.g., TIMEX3~\cite{pustejovsky05slt})—and recognizing relative relations such as BEFORE, AFTER, and OVERLAPS. Temporal reasoning must reconcile document creation time with vague expressions and cross-sentence cues. Second, causal relations~\cite{mirza14etc} capture directed influence (e.g., “The resignation was caused by the scandal”). Signals range from explicit connectives (“because,” “therefore”) to implicit world knowledge, requiring models to handle directionality and complex chains of causes and effects. Third, compositional relations~\cite{araki14dss} address part–whole structures, where sub-events (e.g., “a signing ceremony”) instantiate components of a super-event (e.g., “a peace negotiation”), which supports the creation of hierarchical narratives.

Nevertheless, extracting these event relations remains challenging. It typically requires deep semantic understanding and discourse-level analysis, including long-range dependencies, rhetorical structure, and background knowledge. Moreover, relation extraction interacts bidirectionally with earlier stages: reliable triggers, arguments, and coreference substantially improve relation inference, while relational constraints can, in turn, regularize and correct lower-level decisions.

\subsection{Multimodal EE }
\begin{figure*}[!t]
\centering
\includegraphics[width=\textwidth]{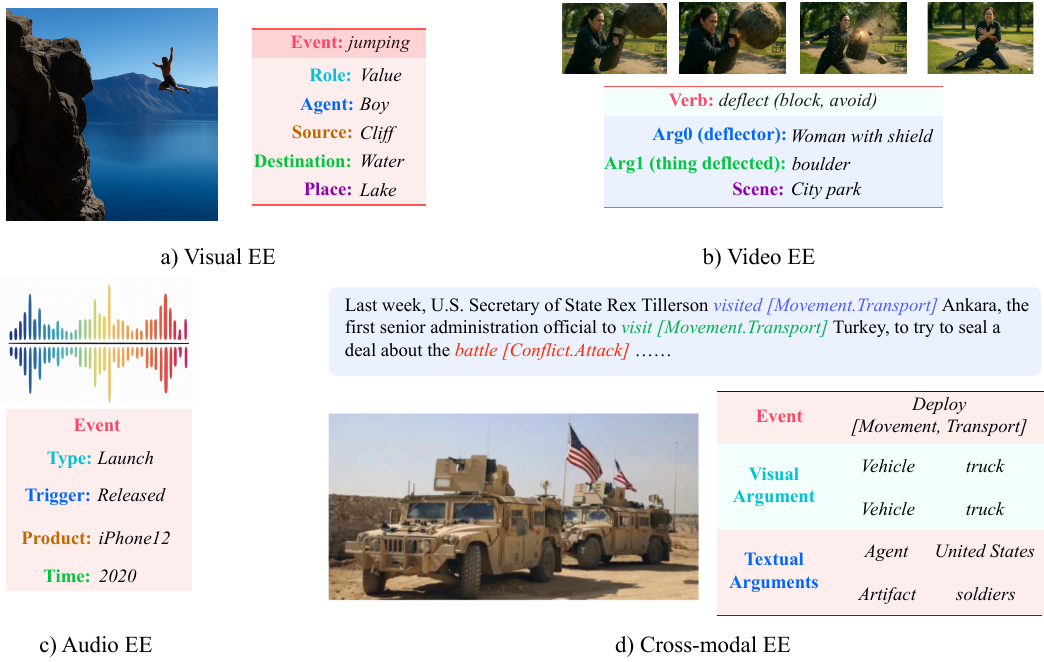}
\caption{Task illustration for different multimodal EE tasks.} 
\label{fig:mmtask}
\end{figure*}

The multimodal EE tasks can be divided into four categories: Visual EE, Video EE, Audio/Speech EE, and Cross-modal EE, as shown in Fig. \ref{fig:mmtask}.
\begin{highlightbox}{Key Difference: Linguistic Context vs. Grounding}
\begin{compactlist}
    \item Text-based EE relies on linguistic cues and discourse context to disambiguate meaning.
    \item Multimodal EE requires grounding, which aligns symbolic roles with physical regions or temporal segments.
\end{compactlist}
\end{highlightbox}

\subsubsection{Visual EE}
Extract events and their semantic roles directly from still images, without relying on accompanying text \cite{yatskar16srv}. This task, also known as situation recognition or visual semantic role labeling, requires predicting both the event trigger (verb) (e.g., cutting, riding) and the associated roles (e.g., agent, tool, object, place), grounding each role to specific visual regions. For instance, given an image of a woman cutting a tomato with a knife in a kitchen, the system should identify the event as cutting, with agent=woman, tool=knife, item=tomato, and place=kitchen. 

\subsubsection{Video EE}  The task of Video EE is to detect and extract events from a given video clip.  We follow the definition of \cite{yang23vee}. 
Given an input video sequence $V = \{ f_i \}_{i=1}^F$ with $F$ frames, the objective of Video EE is to automatically construct 
a set of structured events $\mathcal{E} = \{ e_1, e_2, \dots, e_m \}$ that describe the actions and their participants within the video.  Formally, each event $e \in \mathcal{E}$ is represented as, 
\[
e = \big( v, \langle r^0, a^0 \rangle, \langle r^1, a^1 \rangle, \dots \big),
\]
where $v \in \mathcal{V}$ is an action predicate selected from a predefined verb set $\mathcal{V}$, and each pair $\langle r^k, a^k \rangle$ associates a semantic role $r^k \in \mathcal{R}(v)$  with a corresponding argument $a^k$. The role set $\mathcal{R}(v)$ defines the expected participants for the action $v$ (e.g., \textsc{Agent}, \textsc{Target}, \textsc{Location}, etc.). 
For instance, if the action verb is \textit{knock}, a possible extracted event could be:
$\langle \textsc{Agent}, \text{gray bull} \rangle$, $\langle \textsc{Target}, \text{person in gray hoodie} \rangle$, and $\langle \textsc{Place}, \text{ground} \rangle$.
In this way, Video EE transforms raw video into a set of symbolic event representations, enabling downstream reasoning about ``who did what, to whom, and where'' in the visual domain.

\subsubsection{Audio/Speech EE}
Following the definition of \cite{wang24snb}, we define the Audio/Speech EE task as: given a speech signal represented as a sequence of acoustic frames 
$S = (f_1, f_2, \dots, f_U)$, the goal is to identify and structure events expressed within the audio stream. 
We assume a predefined set of event types $\mathcal{E}$ and a corresponding set of argument roles $\mathcal{R}$. 
Each extracted event record is represented with four components: 
1) an event type $\epsilon \in \mathcal{E}$, 
2) the corresponding event trigger in the audio (e.g., a word or phrase aligned with speech), 
3) an argument role $r \in \mathcal{R}$, 
and 4) the event argument itself. 
Compared with text-based event extraction, Audio/Speech EE faces unique challenges such as noisy signals, disfluencies, speaker variability, and the necessity to align acoustic features with semantic roles. The task therefore requires both robust speech recognition and accurate mapping from speech content to structured event representations.

\subsubsection{Cross-modal EE} 
Cross-modal EE aims to extract structured events by jointly leveraging information from multiple modalities, such as text, vision, and audio \cite{li20cross}.  
Formally, given multimodal inputs $M = \{m^{(t)}, m^{(r)}\}$, where $m^{(t)}$ is the text and $m^{(r)}$ corresponds to another modality (i.e., image, video, or speech), the goal is to identify event types $\epsilon \in \mathcal{E}$, event triggers, and argument-role pairs $\langle r, a \rangle$ across modalities. 
Different modalities provide complementary evidence: for instance, visual cues may reveal objects and actions, text can supply explicit semantic triggers, and audio can indicate speaker roles or affective states.  
The key challenges in Cross-modal EE include aligning heterogeneous signals, grounding arguments consistently across modalities, and reasoning about missing or conflicting evidence. 
By fusing multimodal information, Cross-modal EE can improve event typing accuracy, enhance argument grounding, and support more robust inference compared to unimodal approaches.

\begin{highlightbox}{Key Distinction: Scope \& Boundary}
\begin{compactlist}
    \item Event Extraction (EE) focuses on high-level, schema-defined occurrences with specific semantic roles.
    \item It is distinct from general semantic parsing (like SRL/AMR) and low-level signal processing tasks.
\end{compactlist}
\end{highlightbox}

\section{Methodology}
Event Extraction (EE) systems fundamentally rely on encoding methods to transform unstructured text into structured representations that are amenable to computation. The evolution of these methods reflects the broader history of Natural Language Processing (NLP), progressing from manually crafted linguistic rules to sophisticated, data-driven deep learning architectures. This section provides a comprehensive overview of this evolution, categorized into four major paradigms: rule-based approaches, traditional machine learning, deep learning, and approaches based on large language models (LLMs). Each paradigm offers a distinct approach to capturing the lexical, syntactic, and semantic features necessary to identify event triggers and their corresponding arguments.
To provide a quantitative comparison, we summarize the performance of representative methods across these paradigms in Table \ref{tab:ee-result}.

\subsection{Rule-based Methods}
The foundational paradigm for event extraction was built upon rule-based systems. 
As shown in Fig.~\ref{fig:ee-methods}a, these approaches leverage handcrafted patterns, linguistic heuristics, and syntactic structures to identify event components. Characterized by high precision and interpretability, they often require significant domain expertise and manual effort, which can limit their recall and adaptability to new domains. The development in this area shows a clear trajectory from domain-specific, syntax-heavy systems towards more generalized frameworks and hybrid models that combine rules with statistical methods.

Rule-based methods represent the foundational approach to event extraction, relying on manually crafted patterns and heuristics. Early work in the biomedical domain by \cite{kilicoglu09sdh} utilized syntactic dependency heuristics to identify biological events. This dependency-based paradigm was further explored by \cite{sun11ree}, which framed event extraction as a dependency parsing task. As the field matured, the focus shifted towards creating more generalised and adaptable rule-based systems. For instance, Dutkiewicz et al. \cite{dutkiewicz14rre} introduced a system that could automatically learn rules, reducing the need for extensive manual effort. Similarly, Valenzuela et al. \cite{valenzuelaesca15dir} developed a framework applicable across various domains by abstracting event structures. The application of rule-based methods also expanded to diverse areas like socio-political news analysis \cite{danilova14spe} and artificial social intelligence \cite{nitschke22ree}. To overcome the inherent rigidity of purely rule-based systems, hybrid models emerged. These models combine rule-based components with machine learning to improve performance and adaptability. Kova et al. \cite{kovavc13crm} demonstrated the efficacy of such a hybrid approach for extracting information from clinical texts. This trend was continued by later research, which integrated rules with machine learning for general text \cite{guda17hme, gao19eer}, showcasing the enduring relevance of structured linguistic knowledge in the machine learning era. Most recently, Guda et al. \cite{guda16ree} continued to refine the core rule-based extraction techniques from natural language.

 \subsection{Classic ML Methods}
\begin{figure*}[!t]
\centering
\includegraphics[width=1\textwidth]{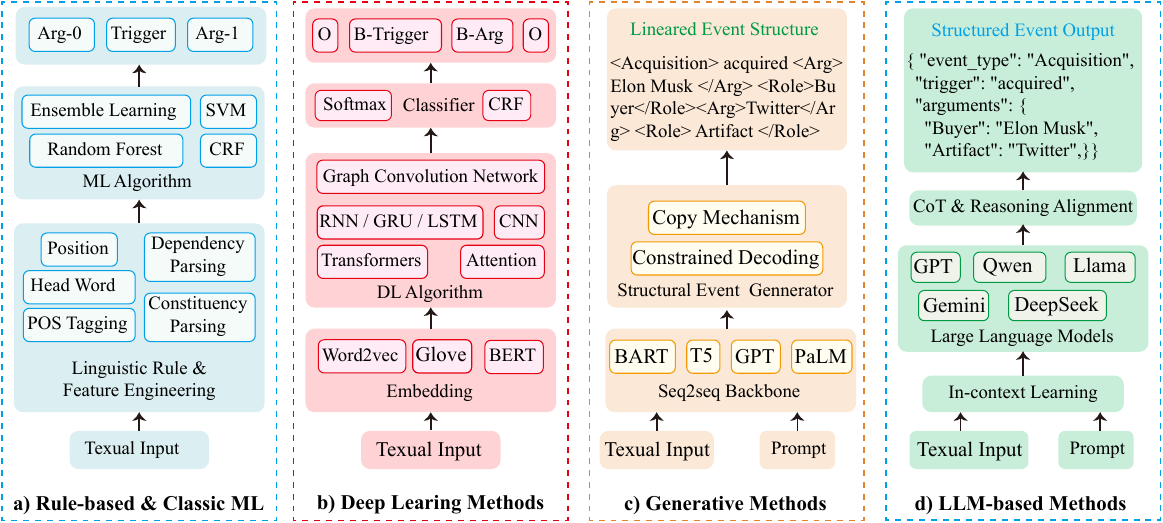}
\caption{Different Methods for Event Extraction.} 
\label{fig:ee-methods}
\end{figure*}

The machine learning era for event extraction shifted the focus from manually crafting rules to training statistical models on annotated data.
As presented in Fig.~\ref{fig:ee-methods}a, this paradigm was characterized by intensive feature engineering, where researchers designed a rich variety of lexical, syntactic, and semantic features to feed into classifiers such as Support Vector Machines (SVMs) and Maximum Entropy models.
A key evolutionary trend during this period was the recognition that sentence-level information is often insufficient, leading to innovative methods that incorporated document-level context to resolve local ambiguities.

Early work established the viability of treating event extraction as a series of classification subtasks. Ahn et al. \cite{ahn06see} conceptualized the task as a pipeline of stages, including anchor identification and argument identification, and used machine-learned classifiers with a rich feature set derived from lexical information and syntactic parsers. Naughton et al. \cite{naughton06eeh} focused on merging event descriptions from multiple news sources, developing methods to group sentences referring to the same event. To improve performance, researchers began incorporating wider contexts.
For example, Ji et al. \cite{ji08ree} and Liao et al. \cite{liao10dlc} both leveraged cross-document inference to refine event extraction. Concurrently, efforts were made to apply these techniques to real-time systems for crisis monitoring \cite{tanev08rtn} and social media analysis \cite{sakaki12rte}.

The focus then shifted towards richer feature engineering and sophisticated modeling.
Patwardhan et al. \cite{patwardhan09ump} developed a unified model incorporating both phrasal and sentential evidence. Miwa et al. \cite{miwa10eec} demonstrated the benefits of using a rich feature set for complex event classification. Hong et al. \cite{hong11cei} proposed cross-entity inference, a method motivated by the observation that entities of the same type often participate in similar events. These methods were successfully adapted to specialized domains such as clinical text \cite{li10lhp}, biomedical literature \cite{bj11gbe}, and social media for detecting adverse drug events \cite{peng16ead}. More recently, machine learning approaches have been applied to new languages like Arabic \cite{smadi18sml} and enhanced with visualization techniques to improve model interpretability \cite{henn21vte}.

\subsection{Deep Learning Methods }
As illustrated in Fig.~\ref{fig:ee-methods}b, The advent of deep learning marked a pivotal paradigm shift in event extraction, moving from manual feature engineering to automatic representation learning.
Neural networks, with their ability to learn hierarchical and distributed features from data, began to systematically outperform traditional machine learning models.
This section chronicles the progression of deep learning architectures, starting with Convolutional and Recurrent Neural Networks, the revolutionary impact of Transformer-based pre-trained language models, and the explicit modeling of syntax with Graph Neural Networks.

\subsubsection{CNN-based Methods}

CNN-based models were among the first deep learning approaches for event extraction, effective at capturing local n-gram features around potential triggers. Nguyen et al. \cite{nguyen15edd} applied CNNs for event trigger detection and showed that a CNN trained on one domain can be adapted to another via domain adaptation techniques. By leveraging pre-trained embeddings and fine-tuning on a target domain, their model achieved robust event detection performance across newswire and novel domains, highlighting CNNs’ ability to learn transferable feature representations. Further improving trigger and argument representation, Chen et al. \cite{chen15eed} introduced a Dynamic Multi-Pooling CNN (DMCNN) that segments the sentence by event arguments and applies multiple pooling operations. This architecture captures features for different sentence segments (before, between, after arguments) separately, which yielded state-of-the-art results by more precisely localizing informative context for each argument role. To jointly extract triggers and arguments, Zhang et al. \cite{zhang16jee} proposed a skip-window CNN that can handle non-contiguous context. Their model uses wider convolutional windows that skip certain distances to capture global sentence features, and it extracts triggers and arguments simultaneously. By expanding the receptive field in the convolution, this joint model better grasped dependencies between distant trigger-argument pairs, improving both trigger classification and argument role labeling. CNNs were also successful in specialized domains: Li et al. \cite{li18ebe} developed a Parallel Multi-Pooling CNN (PMCNN) for biomedical event extraction. PMCNN applies multiple pooling layers in parallel on different parts of the dependency tree or sentence, thereby capturing diverse semantic features (e.g., one pool focusing on trigger context, another on argument context). This parallel design improved extraction of complex biomedical events (like protein modifications) by combining information from multiple context windows. Finally, researchers explored combining CNNs with iterative self-training. Kodelja et al. \cite{kodelja19emg} integrated a CNN-based trigger detector with a bootstrapping approach to incorporate global context. They iteratively refined a global context representation (aggregated from the document or related documents) and fed it into the CNN, allowing the model to use broader topical information. This bootstrapped CNN achieved higher precision on event detection by correcting errors using the larger context (for example, reinforcing that if earlier sentences mention an earthquake, a later sentence’s “magnitude” likely relates to that event).

\subsubsection{RNN-based Methods}

RNNs, particularly Long Short-Term Memory (LSTM) and Gated Recurrent Unit (GRU) networks, became a dominant architecture due to their natural ability to model the sequential nature of text and capture long-range dependencies. Nguyen et al. \cite{nguyen16jee} proposed a joint framework using bidirectional RNNs to predict event triggers and arguments simultaneously, mitigating error propagation by using novel memory features. Chen et al. \cite{chen16eeb} introduced a framework using bidirectional LSTMs with a dynamic multi-pooling layer and a tensor layer to explore the interaction between candidate arguments and predict them jointly.

To address the limitations of local context, Duan et al. \cite{duan17edl} developed a document-level RNN (DLRNN) model capable of automatically extracting cross-sentence clues to improve sentence-level event detection. Researchers also integrated explicit syntactic information into RNNs. Sha et al. \cite{sha18jee} enhanced the RNN architecture with ``dependency bridges'' to carry syntactically related information when modeling each word. Similarly, Li et al. \cite{li19bee} introduced a knowledge-driven Tree-LSTM framework for biomedical event extraction, which explicitly encoded dependency structures and external knowledge from ontologies. To improve feature learning, Zhang et al. \cite{zhang19eed} proposed a multi-task learning framework that incorporated a bidirectional neural language model into the event detection model, allowing it to extract more general patterns from raw data.

\subsubsection{Transformer-based Methods}

Transformer architectures, with their self-attention mechanism and ability to capture long-range dependencies, have been widely adopted for event extraction, especially as they can be pre-trained on large corpora and fine-tuned for extraction tasks. Early explorations integrated transformers with event structures. Yang et al. \cite{yang19ept} examined how pre-trained transformers (like BERT) could be leveraged for event extraction. They proposed separating argument role prediction by roles to handle overlapping arguments and also experimented with using the model to generate event descriptions from structured data. Wadden et al. \cite{wadden19ere} proposed DyGIE++, a unified multi-task framework using contextualized span representations to jointly perform entity, relation, and event extraction. The generative capabilities of Transformer-based models also enabled new task formulations. For instance, Zheng et al. \cite{zheng19dee} introduced an end-to-end framework that generates an entity-based directed acyclic graph, and Lu et al. \cite{lu21tcs} proposed a paradigm that directly generates structured event records from text. To further improve performance, researchers integrated syntactic knowledge into Transformers, as seen in the GATE model \cite{ahmad21gga}, or designed relation-augmented attention mechanisms like RAAT to model argument dependencies in document-level extraction \cite{liang22rra}. Finally, Scaboro et al. \cite{scaboro23eet} provided a valuable benchmark by conducting an extensive evaluation of various Transformer architectures for extracting adverse drug events.

Building on the Transformer, BERT and its variants have been specifically adapted for event extraction. A key innovation was reframing the task itself, such as recasting it as a machine reading comprehension (MRC) problem where a BERT-based model answers questions to identify event details \cite{liu20eem}. To improve performance, researchers developed specialized frameworks like CasEE, which uses a cascade decoding strategy to handle complex overlapping events \cite{sheng21cjl}. Other works focused on adapting the BERT architecture itself, for example, by modifying the self-attention mechanism with mask matrices for Chinese text in MABERT \cite{ding23mma}, or by explicitly integrating event schema annotations into the model input in EABERT \cite{xi23eea}. These advanced models have been successfully applied to critical real-world scenarios, such as extracting information related to COVID-19 \cite{min21eee}, and have been adapted for new languages like Arabic through the creation of dedicated corpora and novel modeling approaches \cite{aljabari24eae}.

\subsubsection{Graph-based Methods}

Graph neural networks (GNNs) have been leveraged to capture structured information (like dependency trees or event graphs) for event extraction. These models propagate node representations through edges, which is intuitive for modeling relationships between triggers and arguments.
Nguyen et al. \cite{nguyen18gcn} first applied Graph Convolutional Networks (GCNs) to event detection. This approach modelled syntactic dependencies as graphs and introduced argument-aware pooling to emphasise potential argument words, improving trigger classification by leveraging dependency structures and argument clues. Liu et al. \cite{liu18jme} developed a model that performs multi-order graph convolution, meaning it not only considers direct neighbors in the dependency graph but also second-order and third-order connections.

Subsequent work refined the GNN approach by incorporating more complex structural information. Yan et al. \cite{yan19edm} extended GCNs to model and aggregate multi-order syntactic representations, moving beyond just first-order dependency relations. Lai et al. \cite{lai20edg} introduced a novel gating mechanism into GCNs to filter noisy information based on the trigger candidate and incorporated syntactic importance scores. Balali et al. \cite{balali20jee} proposed a joint framework that applies GCNs along the shortest dependency path between words, eliminating irrelevant context. Ahmad et al. \cite{ahmad21gga} introduced GATE, a Graph Attention Transformer Encoder that fuses structural information from dependency parses into a self-attention mechanism to improve cross-lingual event extraction. More recently, Wan et al. \cite{wan23mch} built a multi-channel GAT that processes multiple types of relations (dependency, co-reference, etc.) in separate channels and then hierarchically combines them. This approach effectively handles open-event extraction by not assuming predefined event types and by drawing together various relational cues through graph attention. Zhang et al. \cite{wan24mgn} proposed a multifocal graph-based scheme for extracting topic events, essentially identifying a representative event that characterizes a topic from a set of documents.  

\begin{table*}[htbp]
\centering
\caption{Performance comparison of several representative methods on Event Extraction tasks.}
\resizebox{\textwidth}{!}{%
\begin{tabular}{c|l|ccc|ccc|ccc|ccc}
\toprule
\multirow{2}{*}{\textbf{Category}} & \multirow{2}{*}{\textbf{Model}} & \multicolumn{3}{c|}{\textbf{Trigger Id (\%)}} & \multicolumn{3}{c|}{\textbf{Trigger Cls (\%)}} & \multicolumn{3}{c|}{\textbf{Argument Id (\%)}} & \multicolumn{3}{c}{\textbf{Argument Cls (\%)}} \\
\cmidrule(lr){3-5} \cmidrule(lr){6-8} \cmidrule(lr){9-11} \cmidrule(lr){12-14}
 & & \textbf{P} & \textbf{R} & \textbf{F1} & \textbf{P} & \textbf{R} & \textbf{F1} & \textbf{P} & \textbf{R} & \textbf{F1} & \textbf{P} & \textbf{R} & \textbf{F1} \\
\midrule
\multirow{4}{*}{\shortstack{Classic\\ML}}
 & Cross-Event \cite{liao10dlc} & - & - & - & 68.7 & 68.9 & 68.8 & 50.9 & 49.7 & 50.3 & 45.1 & 44.1 & 44.6 \\
 & Cross-Entity \cite{hong11cei}  & - & - & - & 72.9 & 64.3 & 68.3 & 53.4 & 52.9 & 53.1 & 51.6 & 45.5 & 48.3 \\
 & JointBeam \cite{li13jee} & 76.9 & 65.0 & 70.4 & 73.7 & 62.3 & 67.5 & 69.8 & 47.9 & 56.8 & 64.7 & 44.4 & 52.7 \\
 & PSL \cite{liu16psl} & - & - & - & 75.3 & 64.4 & 69.4 & - & - & - & - & - & - \\
\midrule
\multirow{6}{*}{\shortstack{Deep\\Learning}}
 & DMCNN \cite{chen15eed} & 80.4 & 67.7 & 73.5 & 75.6 & 63.6 & 69.1 & 68.8 & 51.9 & 59.1 & 62.2 & 46.9 & 53.5 \\
 & JRNN \cite{nguyen16jee} & 68.5 & 75.7 & 71.9 & 66.0 & 73.0 & 69.3 & 61.4 & 64.2 & 62.8 & 54.2 & 56.7 & 55.4 \\
 & dbRNN \cite{sha18jee} & - & - & - & 74.1 & 69.8 & 71.9 & 71.3 & 64.5 & 67.7 & 66.2 & 52.8 & 58.7 \\
 & ANN-FN \cite{liu16lfi} & - & - & - & 79.5 & 60.7 & 68.8 & - & - & - & - & - & - \\
 & ANN-AugATT \cite{liu17eai} & - & - & - & 78.0 & 66.3 & 71.7 & - & - & - & - & - & - \\
 & JMEE \cite{liu18jme} & 80.2 & 72.1 & 75.9 & 76.3 & 71.3 & 73.7 & 71.4 & \textbf{65.6} & 68.4 & 66.8 & 54.9 & 60.3 \\
\midrule
\multirow{5}{*}{\shortstack{Pretrained \\ Language \\ Model}}
& PLMEE \cite{yang19ept} & 84.8 & 83.7 & 84.2 & 81.0 & 80.4 & 80.7 & 71.4 & 60.1 & 65.3 & 62.3 & 54.2 & 58.0 \\
 & Joint3EE \cite{nguyen19oan} & 70.5 & 74.5 & 72.5 & 68.0 & 71.8 & 69.8 & - & - & - & 52.1 & 52.1 & 52.1 \\
 & GAIL-ELMo \cite{zhang19gail}& 76.8 & 71.2 & 73.9 & 74.8 & 69.4 & 72.0 & 63.3 & 48.7 & 55.1 & 61.6 & 45.7 & 52.4 \\
 & DYGIE++ \cite{wadden19ere} & - & - & - & - & - & 69.7 & - & - & 55.4 & - & - & 52.5 \\
 & BERT-QA \cite{du20dle} & 74.3 & 77.4 & 75.8 & 71.1 & 73.7 & 72.4 & 58.9 & 52.1 & 55.3 & 56.8 & 50.2 & 53.3 \\
\bottomrule
\end{tabular}%
}
\label{tab:ee-result}
\end{table*}

\subsection{LLM-based Methods} 
Current methodologies leveraging LLMs for event extraction can be broadly categorized into six paradigms: \textbf{Instruction Tuning}, \textbf{In-context Learning (ICL)}, \textbf{Chain-of-Thought (CoT)} reasoning, \textbf{Data Augmentation}, \textbf{Multi-agent} frameworks, and \textbf{Multimodal LLMs}.
A schematic overview of these approaches is illustrated in Fig.~\ref{fig:ee-methods}d.
\begin{highlightbox}{Key Difference: Representation vs. Instruction}
\begin{compactlist}
    \item Deep Learning approaches focus on learning complex feature representations to map inputs to labels.
    \item LLM-based approaches emphasize instruction following and reasoning to synthesize structures directly.
\end{compactlist}
\end{highlightbox}

\subsubsection{Instruction-tuned Models.}
Instruction-tuned approaches adapt LLMs to specific event extraction tasks and schemas by directly training them to follow task instructions, thereby providing a promising and cost-effective solution. 
For example, Wang et al. \cite{wang22d} proposed DeepStruct, which unifies diverse structure prediction tasks into triple generation via structure pretraining. Trained on task-agnostic and multi-task corpora, it enhances structural understanding and supports zero-shot and multi-task transfer. 
Wang et al. \cite{wang23imt} proposed InstructUIE, a unified information extraction framework using instruction tuning that reformulates IE tasks in a text-to-text format. Evaluated on the expert-written IE INSTRUCTIONS benchmark, it matches BERT in supervised settings and surpasses GPT-3.5 and prior SOTA in zero-shot performance. 
Li et al. \cite{li24k} proposed KnowCoder, which converts IE schemas into Python-style class representations and uses code pre-training with instruction tuning to improve schema understanding and enable universal information extraction. 
Wei et al. \cite{yuan24back} proposed the first explainable complex temporal reasoning task for predicting future event times with reasoning explanations. They built the 26K-sample ExpTime dataset and released TimeLlaMA, achieving state-of-the-art results in event prediction and explanation generation. 
Sainz et al. \cite{sainz24gl} proposed GoLLIE, which fine-tunes large language models to follow complex annotation guidelines, thereby significantly improving performance on zero-shot information extraction tasks.

Furthermore, recent studies have introduced reinforcement learning and specific optimization strategies to further enhance extraction capabilities.
Qi et al. \cite{qi24a} introduced ADELIE, an IE-aligned large language model trained with the high-quality IEInstruct dataset via instruction tuning and DPO. Its variants, ADELIESFT and ADELIEDPO, achieve state-of-the-art performance across multiple IE benchmarks. 
Xu et al. \cite{wang23c} proposed ChatUIE, a unified information extraction framework based on ChatGLM, which integrates reinforcement learning and generation constraints to effectively improve information extraction performance. 
Hong et al. \cite{hong24bqg1} proposed a reinforcement learning–based question generation method, RLQG, which leverages four evaluation criteria to produce high-quality, generalizable, and context-dependent questions, thereby enhancing the performance of QA-based event extraction. 
Zhang et al. \cite{zhang24u} proposed ULTRA, a framework that improves event argument extraction by hierarchically reading document segments and refining candidate sets, while incorporating LEAFER to better locate argument boundaries. 
Cai et al. \cite{cai24ied} built DivED, an automatically generated dataset with diverse event types and definitions to improve models’ understanding and zero-shot event detection. Fine-tuning LLaMA-2-7B on DivED yields significant gains over traditional methods and even surpasses GPT-3.5 on three open benchmarks. 
Hu et al. \cite{hu25llm} proposed LLMERE, which reformulates Event Relation Extraction as a QA task. It extracts all related events simultaneously and uses partitioning with rationale generation to enhance efficiency, coverage, and interpretability. 
Srivastava et al. \cite{srivastava25itl} integrated human- and machine-generated event annotation guidelines into LLMs for event extraction, improving performance in both full-data and low-resource settings, especially for cross-schema generalization and low-frequency event types.

\begin{table*}[t]
\centering
\caption{Performance comparison on Visual and Multimodal settings. The tasks include Event Trigger Extraction and Argument Role Extraction. Best results are bolded.}
\rowcolors{4}{gray!10}{white} 
\resizebox{\textwidth}{!}{%
\begin{tabular}{lcccccccccccc}
\toprule
\multirow{3}{*}{\textbf{Method}} & \multicolumn{6}{c}{\textbf{Visual (Image-only)}} & \multicolumn{6}{c}{\textbf{Multimodal}} \\
\cmidrule(lr){2-7} \cmidrule(lr){8-13}
 & \multicolumn{3}{c}{Trigger} & \multicolumn{3}{c}{Argument} & \multicolumn{3}{c}{Trigger} & \multicolumn{3}{c}{Argument} \\
\cmidrule(lr){2-4} \cmidrule(lr){5-7} \cmidrule(lr){8-10} \cmidrule(lr){11-13}
 & P & R & F1 & P & R & F1 & P & R & F1 & P & R & F1 \\
\midrule
Flat~\cite{li20cross} & 27.1 & 57.3 & 36.7 & 4.3 & 8.9 & 5.8 & 33.9 & 59.8 & 42.2 & 12.9 & 17.6 & 14.9 \\
WASE~\cite{li20cross} & 43.1 & 59.2 & 49.9 & 14.5 & 10.1 & 11.9 & 43.0 & 62.1 & 50.8 & 19.5 & 18.9 & 19.2 \\
CLIP-EVENT~\cite{li22clip} & 41.3 & \textbf{72.8} & 52.7 & 21.1 & 13.1 & 17.1 & - & - & - & - & - & - \\
UniCL~\cite{liu22mee} & 54.6 & 60.9 & 57.6 & 16.9 & 13.8 & 15.2 & 44.1 & 67.7 & 53.4 & 24.3 & 22.6 & 23.4 \\
CAMEL~\cite{du23tme} & 52.1 & 66.8 & 58.5 & 21.4 & 28.4 & 24.4 & 55.6 & 59.5 & 57.5 & 31.4 & 35.1 & 33.2 \\
UMIE~\cite{sun24uum} & - & - & - & - & - & - & - & - & 62.1 & - & - & 24.5 \\
MMUTF~\cite{seeberger24mmm} & 55.1 & 59.1 & 57.0 & 23.6 & 18.8 & 20.9 & 47.9 & 63.4 & 54.6 & 39.9 & 20.8 & 27.4 \\
X-MTL~\cite{cao25cmm} & \textbf{73.1} & 70.3 & \textbf{71.7} & \textbf{33.2} & \textbf{31.3} & \textbf{32.2} & 78.3 & 57.3 & 66.2 & 40.3 & 42.6 & 41.4 \\
Qwen2VL-7B~\cite{qwen2vl} & 69.11 & 65.38 & 64.83 & 27.30 & 27.63 & 27.31 & 83.77 & 73.14 & 70.64 & 47.13 & 42.70 & 43.04 \\
MKES~\cite{yu25mee} & 69.32 & 65.77 & 65.30 & 27.30 & 27.63 & 27.31 & \textbf{89.82} & \textbf{87.70} & \textbf{87.39} & \textbf{48.11} & \textbf{43.89} & \textbf{44.25} \\
\bottomrule
\end{tabular}%
}
\label{tab:mm_results}
\end{table*}

\subsubsection{In-context Learning.}
ICL-based approaches rely on providing a few-shot context within prompts, enabling LLMs to infer structured information without explicit parameter updates. 
For example, Li et al. \cite{li23eci} systematically evaluated ChatGPT on seven fine-grained IE tasks, finding it underperforms in standard IE but excels in OpenIE and provides high-quality explanations. 
Ma et al. \cite{ma23llm} proposed combining SLM filtering with LLM reranking, using small models for efficient sample filtering and large models for reranking difficult cases, effectively integrating both strengths. 
Pang et al. \cite{pang23glc} proposed Guideline Learning (GL), which automatically generates and retrieves task guidelines to enhance in-context information extraction and mitigate underspecified task descriptions. 
Wang et al. \cite{wang23c} proposed Code4Struct, reformulating structured prediction as code generation. Using event argument extraction, it maps text to class-based event–argument structures, leveraging type annotations and inheritance to integrate external knowledge and constraints. 
Li et al. \cite{wang23c} proposed converting IE outputs into code and using Code-LLMs for code-based extraction, achieving consistent gains over IE-specific and prompt-based LLMs across seven benchmarks.

Complementary to these format-oriented approaches, recent research focuses on optimizing demonstration selection and integrating hybrid workflows to ensure stability and accuracy.
He et al. \cite{he24dra} proposed DRAGEAE, which integrates a knowledge-injected generator with a demonstration retriever to address format inconsistency, multi-argument handling, and contextual deviation in generative event argument extraction. 
Kan et al. \cite{kan24eee} reformulated event extraction as multi-turn dialogues, guiding LLMs to learn event schemas and generate structured outputs. They also introduced an LLM-based data generation method that significantly improves performance on long-tail event types. 
Zhu et al. \cite{zhu24l} proposed LC4EE, which combines SLMs’ event extraction with LLMs’ error correction via auto-generated feedback, allowing LLMs to refine SLM predictions effectively. 
Zhou et al. \cite{zhou24l} proposed HD-LoA prompting for document-level event argument extraction, combining heuristic-driven and analogy-based prompts to help LLMs learn task heuristics from few examples and generalize via analogical reasoning. 
Fu et al. \cite{fu24t} proposed TISE, an example selection method that uses semantic similarity, diversity, and event correlation, applying Determinantal Point Processes to choose contextual examples effectively. 
Bao et al. \cite{bao24gcf} proposed GCIE, a unified information extraction framework that combines LLMs and SLMs in a two-stage pipeline to handle noise, abstract label semantics, and varied span granularity. 
Zhang et al. \cite{zhang24upl} found that large language models exhibit spurious associations in information extraction and proposed two strategies, forward label extension and backward label validation, to leverage the extended labels. 
Wei et al. \cite{wei25hl} proposed CAT (Choose-After-Think), a two-phase “Think–Choose” framework that mitigates LLM preference traps in event argument extraction, greatly enhancing unsupervised EAE. In zero-shot settings, a local 7B model with CAT matches DeepSeek-R1 performance while cutting time costs. 
Liao et al. \cite{liao25r} proposed RUIE, a retrieval-based unified IE framework that combines LLM preferences with a keyword-enhanced reward model and trains its retriever via contrastive learning and knowledge distillation for efficient in-context generalization.

\subsubsection{Chain-of-Thought (CoT).}  

Inject intermediate reasoning into the decoding.  
Improve step-by-step structure induction.

Wei et al. \cite{wei23czs} proposed ChatIE, a prompt-based zero-shot IE method that reformulates extraction as multi-turn QA and leverages ChatGPT across multiple datasets and languages.
Ma et al. \cite{ma24sbl} proposed STAR, which uses LLMs to generate high-quality training data from limited seed examples, boosting performance in low-resource IE tasks like event and relation extraction.
Chen et al. \cite{chen24ill} analyzed LLMs’ weaknesses in event relation reasoning and proposed generative, retrieval-based, and fine-tuning methods with a new dataset, LLM-ERL, significantly improving logical consistency and performance in event relation extraction.
Shuang et al. \cite{shuang24tah} proposed EAESR, a document-level event argument extraction method using guided summarization and reasoning to leverage LLMs’ emergent abilities for key feature extraction and cross-event association.

\subsubsection{Multi-agent Methods}

Wang et al. \cite{wang24doa} proposed DAO (Debate as Optimization), a multi-agent system that refines LLM event extraction outputs through debate without tuning. It integrates DRAG for diverse retrieval and AdaCP for filtering unreliable answers, significantly narrowing the gap between tuning-free and supervised methods on ACE05 and CASIE.
Guan et al. \cite{guan25m} proposed MMD-ERE, a multi-agent debate framework for event relation extraction, where cooperative and confrontational debates with audience feedback enhance relation understanding. It outperforms baselines across multiple ERE tasks and LLMs, demonstrating the effectiveness of the debate mechanism.

\subsubsection{Data Augmentation with LLMs}

Wang et al. \cite{wang24tal} proposed TALOR-EE, which enhances low-resource event extraction through targeted augmentation, negative sampling, and back-validation, achieving notable gains in zero- and few-shot settings.
Jin et al. \cite{jin24sda} proposed a schema-based data augmentation method that leverages event schemas to generate synthetic data, thereby alleviating the scarcity of annotated data in event extraction tasks.
Choudhary et al. \cite{choudhary24q} proposed QAEVENT, a document-level event representation that models events as question–answer pairs, removing predefined role schemas and improving annotation efficiency and coverage.
Zhao et al. \cite{zhao242ii} proposed MoRAG-FD, a biomedical event causality framework using retrieval-augmented multi-perspective data expansion and fine-grained denoising via syntactic dependency–based weighting of irrelevant entity pairs.
Chen et al. \cite{chen24llm} used LLMs as expert annotators to expand event extraction datasets, aligning generated data with benchmark distributions to mitigate data scarcity and imbalance.
Uddin et al. \cite{uddin24guc} proposed automatic question generation methods for document-level event argument extraction, producing both contextualized and uncontextualized questions without human input. Combining the two notably improves performance, especially for cross-sentence triggers and arguments.
Meng et al. \cite{meng24c} proposed CEAN, a contrastive event aggregation network with LLM-based augmentation that uses event semantics and contrastive learning to reduce noise and boost low-resource event extraction.
Zhou et al. \cite{zhou25bdf} proposed BiTer, a bidirectional feature learning method for zero-shot event argument extraction that jointly learns contextual and labeled features while using LLMs to generate pseudo-arguments to reduce context bias and improve feature representation.

\subsubsection{Multimodal LLMs (MLLMs).}  

Extend LLMs with vision or audio inputs.  
Enable cross-modal trigger and role reasoning.

Bao et al. \cite{bao24egi1} proposed a multimodal Chinese event extraction model that incorporates character glyph images to capture intra- and inter-character morphological features.
Ma et al. \cite{ma25sms} proposed the M-VAE task for extracting and localizing abnormal event quadruples in videos. Their Sherlock model, featuring Global-local Spatial-enhanced MoE and Spatial Imbalance Regulator modules, effectively captures spatial information and significantly outperforms existing Video-LLMs on the M-VAE dataset.
Bao et al. \cite{bao25rc} proposed a Literary Vision-Language Model for classical Chinese event extraction that integrates literary annotations, historical context, and glyph features to capture rich semantic information.
Zhang et al. \cite{zhang24rea} proposed the first framework of Multimodal Universal Information Extraction (MUIE) and developed a multimodal large language model, REAMO, which can perform information recognition and fine-grained grounding across various modalities.

We provide a schematic illustration of these multimodal EE methodologies in Fig.~\ref{fig:ee-methods}.

\begin{figure*}[!t]
\centering
\includegraphics[width=1\textwidth]{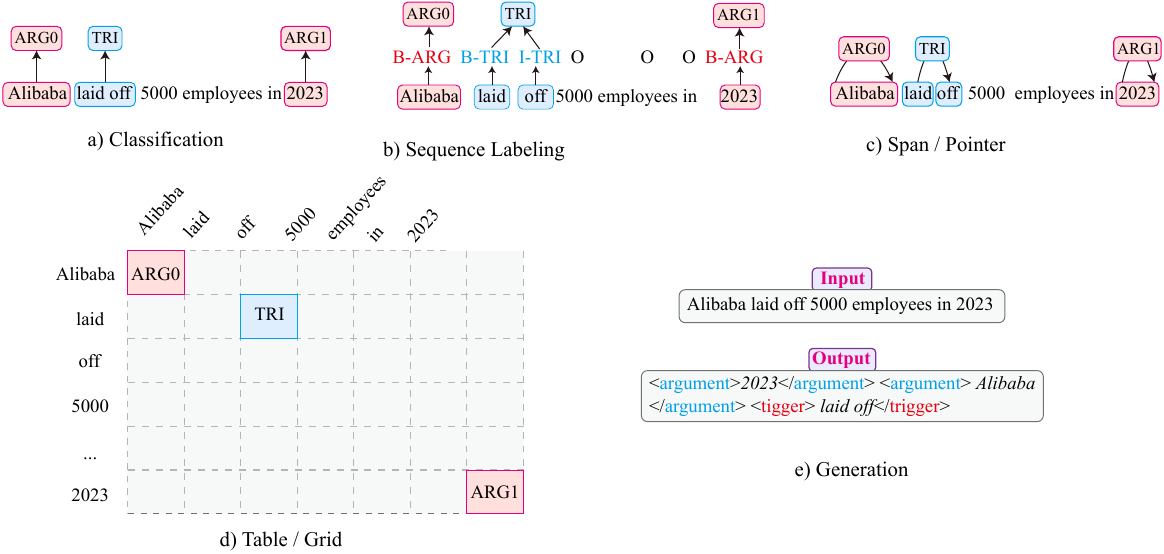}
\caption{Different Decoding Paradigms for Event Extraction.} 
\label{fig:decoding}
\end{figure*}

\section{Formulations \& Decoding }
Event extraction typically involves multiple subtasks, such as trigger word and argument extraction and their type classification, event relationship extraction, and event coreference resolution. Therefore, in this section, we categorize and summarize existing event extraction methods based on their decoding approach, including classification methods, sequence labeling methods, span/pointer methods, table/grid methods, and generation methods. The decoding diagrams of these methods are shown in Fig.~\ref{fig:decoding}.
\subsection{Classification}

Classification methods typically classify the types of entities such as extracted trigger words and arguments, or classify the relations between events. Some existing methods improve model performance by incorporating lexical, syntactic, and semantic knowledge into classification models \cite{nguyen16jee,li13jee,liu18jme,yang16jee,ece16zhao}. For example, Chen et al. \cite{chen12jmc} proposed a knowledge-rich linguistic feature approach that not only effectively utilizes character-level features but also incorporates the results of zero pronoun resolution and noun phrase coreference resolution, as well as features such as trigger word probability and trigger word type consistency. These features capture rich linguistic information from the character to the utterance level. Furthermore, some methods transform trigger words and arguments into graph or tree structures \cite{mcclosky11eed,zheng19dee,wang21ccp,li19bee,huang16lee}. For example, Han et al. \cite{jet19han} utilize structured prediction to simultaneously extract event and temporal relations. They model event relations as a graph structure that includes three types of edges: event-relation consistency, transitivity of temporal relations, and symmetry. Whether incorporating additional knowledge or constructing a graph (tree) structure, the ultimate goal of these methods is to better model trigger words, arguments, and their relations for better classification.

\subsection{Sequence Labeling}
Sequence labeling methods treat event extraction as a sequence labeling problem. Unlike classification methods, sequence labeling methods can simultaneously extract trigger words and arguments, as well as their types. However, such methods have difficulty handling nested or multi-event scenarios. Typically, sequence labeling methods use a labeling system similar to named entity recognition (NER) (such as BIO tags) and use CRF for label prediction. Most existing research uses weak supervision to generate more training data to improve model performance \cite{yang18ddl,zeng18sue,ferguson18sse}. For example, Kan et al. \cite{kan24llf} used queries, internet retrieval, and LLM-based question answering to generate pseudo-labels to form a weakly labeled dataset. They then used the automatically labeled weakly labeled data to pre-train the event extraction model, and finally fine-tuned the pre-trained model on a manually labeled dataset. In addition, He et al. \cite{he14icr} repeatedly embedded the trigger words of candidate events into the gaps between each word in the sentence to form a new extended sequence, thereby strengthening the trigger word's presence in the sequence. Ramponi et al. \cite{ramponi20bee} designed a multi-label-aware encoding strategy that represents the label of each token as a triple and handles nested event structures through relative position encoding. Lu et al. \cite{lu12aee} introduced customizable ``structural preferences'' (such as grammatical rules and event patterns) to guide model learning.

\subsection{Span/Pointer}

Span-based methods extract events by predicting the spans (start and end positions) of trigger words and arguments. These methods can handle nested structures with multiple trigger words or arguments within the same sentence, but their drawback is the need to enumerate all possible span combinations, resulting in low efficiency. Existing span-based methods fall into two main categories. One approach captures global relationships by constructing graph structures to better predict spans \cite{shuang24tah,wan25epi,xu22tsa}.
For example, Yang et al. \cite{yang23alp} compress unrelated subgraphs and edge types, integrates text span information, and highlights surrounding events within the same document. Finally, they identify event arguments by predicting edges between event trigger words and other nodes. The other approach treats event extraction as a question-answering task, predicting spans through questioning \cite{li20eem,wei21tns,liu21mrc,liu20eem,du20eea}. For example, Zhou et al. \cite{zhou21wrv} extract spans through a collaborative question-answering process: identifying arguments by asking, ``What plays [role] in [event type]?'' and identifying roles by asking, ``What role does [argument] play in [event type]?''

\subsection{Table/Grid}
The grid tagging method converts text into a two-dimensional grid to predict events and their relations. The advantage of this method lies in its ability to handle complex scenarios such as nested and multiple events simultaneously, making it more time-efficient than span-based methods. However, since each cell in the grid represents a word-pair relationship, the size of the two-dimensional grid increases exponentially with longer text, consuming a significant amount of space. Cao et al. \cite{cao22oos} pioneered the grid tagging method for event extraction, designing two grids to extract trigger words and arguments, respectively. Cui et al. \cite{cui22ece} later developed a dual-grid annotation scheme to capture correlations between event parameters within and across events, thereby addressing event causality. Pu et al. \cite{pu24jfh} improved on the method of Cui et al. by designing a heterogeneous relational perception graph module and a multi-channel label enhancement module. Ning et al. \cite{ning23oos} initially used a single grid and four vertex labels to simultaneously extract trigger words and arguments. Chen et al. \cite{chen25dgt} later tackled event causality identification using a single grid and six labels. Wan et al. \cite{wan23jdl} incorporated the eType-Role1-Role2 composite labeling and a complete subgraph-based decoding strategy into the grid tagging model to handle more complex document-level event extraction.

\subsection{Generation}

Generative methods use generative models to generate structured event representations directly from text. The advantage of these methods is that they don't require decomposition into subtasks, directly outputting event types, trigger words, and arguments, thus avoiding error propagation between subtasks. They can also handle a wide range of complex events. However, generative methods often suffer from hallucination problems, generating trigger words or arguments that don't exist in the text, or outputting results that don't conform to the structured representation. Existing methods typically construct task-specific prompt templates, generate these templates, and then decode the corresponding results \cite{li21dle,lu21tcs,hsu22dde,liu22dpt,wen21rds, fei22lasuie}. 
With the rapid development of LLMs, LLM-based prompt engineering and instruction tuning techniques are gaining increasing attention among researchers \cite{xu25miu,sun24uum,zhang24uul,zhou24llt,zhang24rea}. Furthermore, some work considers EE as a QA task. In fact, QA can also be considered a form of prompt engineering, essentially still directly generating events by constructing task-specific question and answer templates \cite{du22rag,lu23eeq,hong24bqg}. Furthermore, Wang et al. \cite{wang19oee} first applied generative adversarial networks to open-domain event extraction. GANs generate events by learning the mapping between document-event distributions and event-related word distributions. Ren et al. \cite{ren23rsd} further improved the performance of generative models using retrieval enhancement techniques. They designed three retrieval methods: context-consistent retrieval (retrieval of similar documents in the input space), pattern-consistent retrieval (retrieval of similar tags in the tag space), and adaptive hybrid retrieval.
\begin{highlightbox}{Trade-off: Flexibility vs. Faithfulness}
\begin{compactlist}
    \item Generative models support open schemas and end-to-end extraction without cascading errors.
    \item However, they are prone to hallucination, creating plausible but non-existent details lacking source evidence.
\end{compactlist}
\end{highlightbox}

\begin{figure*}[!t]
\centering
\includegraphics[width=1\textwidth]{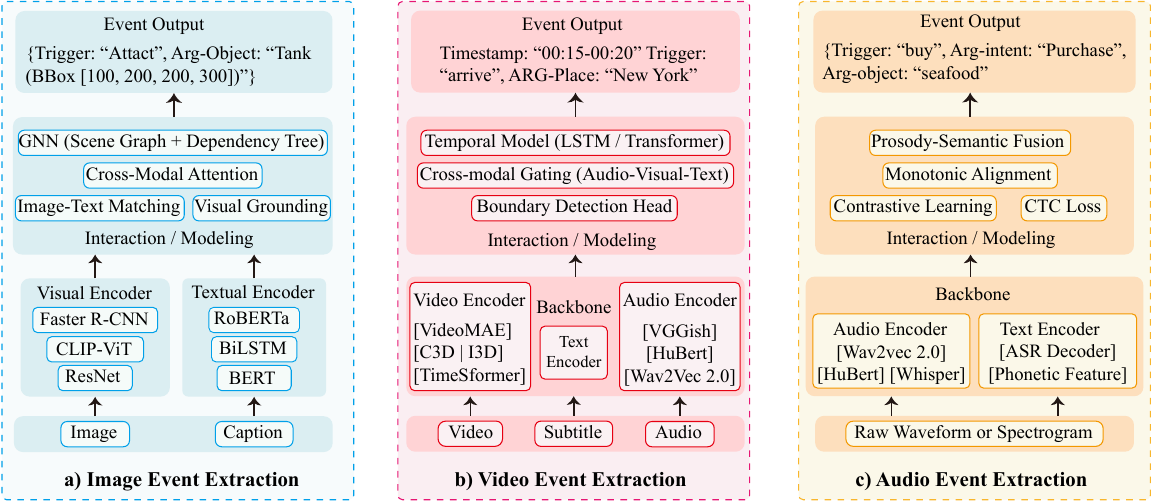}
\caption{Different Methods for Multimodal Event Extraction.} 
\label{fig:mm_methods}
\end{figure*}

\section{System Architectures }

Event extraction is a complex information extraction task that inherently involves multiple interdependent subtasks, including trigger word extraction, event type classification, argument identification, and argument role classification. The relationships between these subtasks are critical, as the identification of an event trigger often dictates the schema for potential arguments. In this section, we categorize existing methodologies based on their architectural design choices. These choices fundamentally determine how the dependencies between subtasks are modeled and how information flows through the system. We classify the architectures into three primary categories: Pipeline, Joint/Global, and One-Stage/Unified models. The schematic overview of these architectures is illustrated in Fig.~\ref{fig:decoding}.

\subsection{Pipeline}
The pipeline architecture represents the classical approach to event extraction, where the problem is decomposed into a series of distinct, sequential subtasks. In this framework, separate models are trained for each stage, and the output of one stage serves as the input for the next. Typically, the process begins with trigger word identification, followed by event type classification, and concludes with argument extraction and role classification \cite{chen15eed,ji08ree,du20eea,liao10dlc,yang19ept}. This modular design offers the advantage of simplicity and interpretability, as each sub-model focuses on a specific, constrained problem. Furthermore, it allows for the flexible combination of different algorithms for different stages.

However, the sequential nature of pipeline models introduces two significant limitations. First, they suffer from severe error propagation cascades. Since downstream models for argument extraction rely entirely on the predictions of upstream trigger classifiers, any error in trigger identification or classification is irreversible and inevitably leads to failures in argument extraction. The downstream models have no mechanism to correct upstream errors. Second, this architecture leads to information fragmentation. By treating subtasks as statistically independent steps, the system fails to leverage the semantic interactions between tasks. For instance, the presence of specific arguments often provides strong contextual clues that can help disambiguate the event type, but a strict pipeline prevents this reverse information flow. 

A representative example is the work of Liu et al. \cite{liu20eem}, who formulated event extraction as a Machine Reading Comprehension (MRC) task. Their system operates in a strict sequence: it first utilizes a special query token [EVENT] to identify triggers and classify event types. Subsequently, it constructs natural language questions based on the predicted types to extract arguments using a BERT-based MRC model. While effective, the dependence is unidirectional; if the initial trigger detection fails, the subsequent question generation relies on incorrect premises, rendering the argument extraction step futile.

\subsection{Joint/Global}
To mitigate the limitations of the pipeline approach, Joint (or Global) architectures model multiple subtasks within a single unified framework. Unlike pipeline methods where modules are trained in isolation, joint models optimize subtasks simultaneously, typically sharing a common encoder or feature representation layer. The core philosophy is that trigger detection and argument extraction are mutually informative; knowing the arguments can clarify the event type, and vice versa. Consequently, these models aim to learn a global representation that captures the dependencies between triggers and arguments \cite{nguyen16jee,li13jee,liu18jme,lin20jnm,sha18jee}.

Joint architectures effectively alleviate the error propagation problem by replacing the hard decisions of upstream pipeline steps with soft, shared parameter optimization. This allows the model to adjust its internal representations based on the loss from all subtasks during training. For example, Graph Neural Networks (GNNs) are frequently employed in this category to explicitly model the structural connections between words. Xu et al. \cite{xu21dle} proposed a sophisticated heterogeneous graph network that integrates sentence nodes and entity mention nodes. Their model captures global interactions through various edge types, including inter-sentence edges and cross-sentence mention edges. By performing message passing over this graph, the model achieves collaborative learning, where the features for entity extraction, event detection, and argument role labeling are jointly refined. This holistic view ensures that the extracted events are structurally consistent and semantically coherent.

\subsection{One-Stage/Unified}
The One-Stage, or Unified, architecture represents a paradigm shift in event extraction, moving away from task-specific classification heads towards a fully end-to-end generation or prediction process. In this approach, the distinction between trigger detection and argument extraction is minimized. Instead of designing separate loss functions or modules for different subtasks, unified methods employ a singular architecture to output the complete event structure directly. These models are predominantly generative, often leveraging the capabilities of Pre-trained Language Models (PLMs) or Large Language Models (LLMs) to transform event extraction into a sequence generation problem \cite{li21dle,zhang24rea,ma22pep,liu22dpt,sun24uum}.

The primary advantage of unified frameworks is their streamlined design and flexibility. By converting the extraction task into a text-to-structure or text-to-text format, these methods avoid the complexity of designing specific network components for each subtask. Furthermore, they are highly adaptable to new event schemas through prompt engineering rather than architectural re-engineering. Researchers can design task-specific prompts that instruct the model to linearize the event structure. For instance, Lin et al. \cite{lin25gge} introduced a multi-perspective prompt design strategy. Their method does not treat event extraction as a labeling task but rather queries the model with various prompt arrangements for each event type. This allows the model to leverage diverse semantic perspectives to understand the context and generate the full event record in a single pass. By optimizing a single objective function, unified models ensure that all components of an event are generated in a globally optimal manner, reducing the disconnect often seen in multi-stage systems.
\begin{highlightbox}{Architecture: Pipeline vs. Unified}
\begin{compactlist}
    \item Pipeline models decompose tasks sequentially, leading to irreversible error propagation.
    \item Unified models treat EE as a single prediction, maximizing information flow across subtasks.
\end{compactlist}
\end{highlightbox}

\section{Representations \& Feature Engineering}
\subsection{Lexicon}
Early EE approaches relied heavily on curated lexical resources such as trigger lists and gazetteers to bootstrap event recognition in the absence of large annotated corpora. These resources~\cite{huang20bee,veyseh20gtn} provided simple but effective anchors for detecting candidate triggers and arguments, often combined with rule-based or pattern-matching systems to enhance recall. As research progressed, lexicons were increasingly integrated into statistical and neural pipelines as weak supervision signals~\cite{lai21eeh}, supplying prior knowledge to guide trigger identification or constrain argument spans. More recent work~\cite{sun23sne,ning23oos} explores hybrid strategies where lexicon-derived cues complement distributed representations, showing that gazetteer augmentation and dictionary-based triggers can still provide robustness under domain shift or low-resource scenarios.  Recent studies~\cite{gao24bee,dukic23loi} continue to highlight the auxiliary role of lexical priors within large-scale or domain-adaptive EE frameworks, revisiting curated trigger word lists as a lightweight yet interpretable supervision source.

\subsection{Syntactic \& Semantic}
Syntactic and semantic structures remain central to event extraction because they encode relational signals that guide how triggers connect to arguments and roles. Early approaches~\cite{veyseh20gtn,abdulkadhar21mlg} integrated POS tags, constituency boundaries, and dependency arcs into neural encoders, showing that structural cues improve argument boundary precision and help reduce error propagation in complex sentences. Graph-based models~\cite{ahmad21gga,ren23tsn} further leveraged syntactic trees to propagate evidence across spans, improving trigger–role interactions and global coherence.

Beyond syntax, SRL and frame semantics provide predicate–argument abstractions that naturally align with event schemas. SRL-based supervision~\cite{zhang22tls,vannguyen22lct} has been used as auxiliary signals, constraints, and multi-task objectives, consistently enhancing role disambiguation and generalization. Several studies~\cite{ding22eri,tao22mre,jiao22ova} design explicit role interaction networks or fuse role semantics with span–relation modeling, further improving robustness in multilingual and biomedical settings. More recently, researchers have tested whether large language models~\cite{chen24llm,menini24sfe} can act as annotators of semantic roles, with mixed results depending on domain specificity.

Abstract meaning representation (AMR) introduces a higher-level structural bias: nodes and edges normalize lexical variation and facilitate reasoning beyond sentence boundaries. Document-level event extraction~\cite{xu22tsa,hsu23aaa,yang23alp} benefits from AMR-enhanced models, including two-stream architectures, prefix injection, and AMR-based link prediction. Dual-level AMR~\cite{huang25dla,wan25epi} injection and event pattern instance graphs further extend this line by embedding AMR into pretrained models for cross-sentence role resolution.

At the discourse level, structural cues such as coreference and narrative flow help connect argument mentions scattered across documents. Systems that explicitly project narrative knowledge or exploit discourse signals~\cite{tang21dnk,li21dle} enhance argument completion and role consistency. Coupling coreference with dependency and SRL information~\cite{du20dle,wen21ete,ren22cri} further enables implicit role reasoning and temporal coherence across events. Beyond explicit discourse modeling, graph-augmented architectures~\cite{li23iei,zhang24smg} capture intra- and inter-event dependencies, highlighting that explicit structure remains indispensable for enforcing discourse-level coherence and role resolution.

Building on these insights, recent research moves from treating syntactic and semantic structures as standalone alternatives to pretrained embeddings toward integrating them as complementary inductive biases. Syntax-guided latent variable models~\cite{zhuang23sdl,hao25ssr}, reinforcement-driven boundary refinement, and soft structural constraints inject syntax into PLMs to improve calibration in long-range argument attachment and low-resource scenarios. Hybrid approaches~\cite{yao25crr} further demonstrate that schema-aware decoding and role relevance reallocation can effectively combine structural signals with contextual embeddings. This line of work~\cite{du20ggr,wang23cee,zhang24sea,fane25bmb} consistently shows that while pretrained embeddings capture rich local semantics, explicit structural modeling is crucial for supporting reliable generalization across both sentence- and document-level extraction.
\subsection{Knowledge Retrieval}
Research on event extraction has increasingly emphasized that local sentence context alone is often insufficient for robust trigger identification or argument filling, particularly when roles are implicit or span multiple sentences. To address this limitation, a diverse body of work has explored the integration of external knowledge sources—including structured ontologies, curated schemas, knowledge graphs, retrieved documents, and commonsense resources—into extraction pipelines. Early studies~\cite{han20dke,shen20hcl,wang20jcl,huang20bee} grounded arguments in domain ontologies or legal schemas, showing that factual constraints help disambiguate types and enforce role consistency. These ideas motivated retrieval-based architectures~\cite{wang21qer,zhang22kem,veyseh22eev,huang21esc} in which supporting passages or entity descriptions are gathered dynamically from external corpora and then fused with contextual encoders via attention or gating, thereby reducing errors in argument completion.

Knowledge graphs~\cite{liang22rra,zhuang23kee} quickly became central in this paradigm: aligning mentions with KG nodes and propagating signals along edges provides schema compatibility, while retrieving prototypical schema instances improves zero-shot transfer. Commonsense KGs such as ConceptNet and ATOMIC have also been employed, with subgraph retrieval~\cite{liu20eem,yang23alp,li23iei}. guiding implicit argument reasoning and improving recall where surface cues are sparse. Retrieval has likewise been coupled with cross-domain and cross-lingual adaptation~\cite{zhang22ezs,popovic22fsd,parekh22gbg}, where external knowledge mitigates data sparsity and enhances generalization in specialized or low-resource scenarios. 

\begin{table*}[t]
\centering
\setlength{\tabcolsep}{6pt}
\renewcommand{\arraystretch}{1.25}
\rowcolors{2}{gray!10}{white}
\caption{Representative event extraction benchmarks. “Scale” is a rough relative indicator (Small/Medium/Large) to help compare dataset sizes at a glance. \textbf{ERE}: Event Relation Extraction; \textbf{ECR}: Event Coreference Resolution.}
\begin{tabular}{%
    >{\raggedright\arraybackslash}p{2.5cm}
    >{\raggedright\arraybackslash}p{2.8cm} % 稍微调宽了这一列以容纳 "Event Extraction"
    >{\raggedright\arraybackslash}p{2.8cm}
    >{\raggedright\arraybackslash}p{1.8cm}
    >{\raggedright\arraybackslash}p{1.8cm}
    >{\raggedright\arraybackslash}p{3.8cm}}
\toprule
\textbf{Name} & \textbf{Task} & \textbf{Domain} & \textbf{Scale} & \textbf{Language} & \textbf{Notes} \\
\midrule
ACE05 \cite{doddington04ace1} & Event Extraction & Newswire, broadcast & Medium & EN / ZH / AR & trigger, argument \\
MUC-4 \cite{sundheim92of} &  Event Extraction & Terrorism newswire & Small & EN & trigger, argument \\
MAVEN \cite{wang20m} &  Event Extraction & Wikipedia & Large & EN & trigger, argument \\
RAMS \cite{ebner20msa} &  Event Extraction & News & Medium & EN & trigger, argument \\
WikiEvents \cite{li21dle} &  Event Extraction & Wikipedia + news & Medium & EN & trigger, argument \\
CASIE \cite{satyapanich20cec} &  Event Extraction & Cybersecurity news & Small–Medium & EN & trigger, argument \\
PHEE \cite{sun22p} &  Event Extraction & Pharmacovigilance & Small–Medium & EN & trigger, argument \\
DuEE \cite{li20dls} &  Event Extraction & Open-domain news & Medium–Large & ZH & trigger, argument \\
DuEE-Fin \cite{han22dfl} &  Event Extraction & Finance news & Medium & ZH & trigger, argument \\
FewEvent \cite{deng20mld} &  Event Extraction & Open-domain & Small–Medium & ZH & trigger, argument \\
\midrule
Rich ERE \cite{song15lre1} & ERE & News, forums & Medium & EN / ZH / AR & trigger, argument, relation \\
MAVEN-ERE \cite{wang20m} & ERE & Wikipedia  & Large & EN & trigger, argument, relation \\
HiEve \cite{glavavs14h} & ERE & News (temporal) & Small–Medium & EN & trigger, argument, relation \\
\midrule
GENIA \cite{kim11o} 
& ECR & Paper& Medium & EN & trigger, argument, coreference \\
MLEE \cite{pyysalo12eea} 
& ECR & Biomedical Paper & Medium & EN & trigger, argument, coreference \\
KBP2017 \cite{choubey17tak} & ECR & News + forums & Medium & EN & trigger, argument, coreference \\
\bottomrule
\end{tabular}
\label{tab:ee-benchmarks}
\end{table*}

More recent~\cite{du22rag,deng222bo,ren23rsd} advances integrate retrieval into generative pipelines. Retrieval-augmented generation (RAG) models combine dense retrievers with generative decoders that construct structured frames under schema constraints, outperforming purely end-to-end baselines in open or evolving domains. Other approaches~\cite{huang23eed,siagian23eet,feng22ljp} retrieve schema descriptors, event exemplars, or temporal patterns that regularize role inventories and support schema-aware decoding. Large-scale pipelines~\cite{zhang23als,kim23ada} automatically build or synthesize event knowledge, offering broader coverage and complementing retrieval modules in emerging domains. In parallel, generative and continual learning frameworks~\cite{du20ggr,ahmad21gga,wang23cee} highlight how retrieval can be integrated with role reasoning and structural constraints for greater robustness.

\subsection{Pretrained Embeddings}

Pretrained language models (PLMs)~\cite{devlin19bpt,liu19rro,joshi20sip} make event extraction largely a feature–engineering problem: contextual token and span representations are harvested from a shared encoder and reused everywhere—trigger scoring, argument identification, and role typing—so that one universal embedding space underpins the entire pipeline. In sentence-level EE, PLM features replace brittle lexical lists and hand-built templates, yielding stronger trigger detectors and tighter argument boundaries, with span pooling or biaffine heads operating directly on contextualized vectors. Domain-tailored PLMs~\cite{yan20cae,su24gbe}. further align specialized terminology with role inventories, improving calibration and recall without bespoke features.

Document-level models~\cite{veyseh20gtn,ahmad21gga,hsu23aaa,huang25dla} keep the same recipe—one encoder, many heads—but add structure-aware aggregation atop PLM features (graph layers, AMR-aware prefixes, role-guided attention) to propagate evidence across mentions while staying in the PLM embedding space. Parameter-efficient conditioning~\cite{ma22pep,liu22dpt,liu22dpt,nguyen23csp,ma23ssp} then shapes these universal features toward schema constraints and label spaces with minimal finetuning, improving few-shot trigger recognition and argument-role generalization. 

More recently, generative adaptations~\cite{lewis19bds,raffel20elt,du22rag,he24dra,uddin24guc} still rely on the same pretrained representation core—now used to produce frames directly under decoding constraints—and can be fused with retrieval or demonstrations when roles are implicit, reinforcing the view that PLM-derived contextual embeddings function as universal features across detection, role typing, and cross-sentence aggregation.

\subsection{Visual Feature}

Visual feature extraction for situation recognition and video event extraction typically combines {local, object-centric cues} with {global, scene-level context}. 
Early works mainly used global CNN embeddings (e.g., VGG-16 \cite{simonyan14vdc} or ResNet-50 \cite{he16drl}) for situation recognition \cite{yatskar16srv,yatskar17cus,mallya17rms,li17srg,suhail19mkg,cooray20aca}, focusing on capturing the holistic scene for verb prediction. Subsequent grounded approaches explicitly incorporated local object detectors such as Faster R-CNN \cite{ren16fcr} and RoI-pooled embeddings to jointly predict role labels and bounding boxes \cite{pratt20gsr,cho21gsr}, while later extensions further exploited transformers and collaborative attention to refine local-global fusion \cite{cho22ctg,yu23kag,wang25eev}. Beyond static images, video-based models rely on both global spatio-temporal encoders (e.g., I3D \cite{carreira17qva}, SlowFast \cite{feichtenhofer19snv}, Video Swin Transformer \cite{liu22vst}) to capture temporal dynamics \cite{sadhu21vsr,gao21vrd,khan22gvs,zhao23chs} and local tracklet-based features that trace entities and their evolving states over time \cite{yang23vee,zhao23chs,liu24mgg,sugandhika25ssg}. More recent works extend these pipelines with structural modeling, such as spatio-temporal scene graphs, visual analogies, and causality-guided attention \cite{suhail19mkg,pratt20gsr,cho21gsr,yu23kag,bitton23vva,chen23vvs,wang25eev}. Together, these efforts highlight a clear trend: local features provide fine-grained entity grounding ~\cite{liu24md} and argument role understanding, while global representations capture holistic context and temporal coherence, and their integration has become the standard backbone for robust visual event extraction.

\section{Dataset \& Evaluation}

\subsection{Text Datasets }

We survey sentence- and document-level datasets for text-based event extraction, covering resources annotated for event detection, event relations, and event coreference.

\begin{table*}[t]
\centering
\rowcolors{2}{gray!10}{white}
\setlength{\tabcolsep}{5pt}
\renewcommand{\arraystretch}{1.25}
\caption{Multimodal event extraction benchmarks.}
\begin{tabular}{%
    >{\raggedright\arraybackslash}m{2.8cm}
    >{\raggedright\arraybackslash}m{2.8cm}
    >{\raggedright\arraybackslash}m{2.8cm}
    >{\raggedright\arraybackslash}m{2.2cm}
    >{\raggedright\arraybackslash}m{1.8cm}
    >{\raggedright\arraybackslash}m{4.0cm}}
\toprule
\textbf{Name} & \textbf{Modality} & \textbf{Domain} & \textbf{Scale} & \textbf{Language} & \textbf{Output} \\
\midrule
imSitu \cite{yatskar16srv} & Image & General & 126,102 & English & Verb, semantic roles and values \\
SWiG \cite{pratt20gsr} & Image & General & 126,102 & English & Verb, semantic roles, values, bounding boxes \\
VASR \cite{bitton23vva} & Image & General & 3,820 & English & Verb, semantic roles and values \\
VidSitu \cite{sadhu21vsr} & Video & Movie & 126,102 & English & Verb and values \\
Grounded VidSitu \cite{khan22gvs} & Video & Movie & 126,102 & English & Verb, values, bounding boxes \\
SpeechEE \cite{wang24snb} & Audio & News, medical, biology, cybersecurity, movies & 5,240 human + 49,585 synth & EN / ZH & Event type, trigger, argument role \\
M$^2$E$^2$ \cite{li20cross} & Text + Image & News & 245 & English & Event type, trigger, image object \\
VOANews \cite{shih22cec} & Text + Image & News & 106,875 & English & Event type, trigger, image object \\
CMMEvent \cite{liu25cme} & Text + Image & News & 5,709 & Chinese & Event type, trigger, role \\
VM$^2$E$^2$ \cite{chen21jme} & Text + Video & News & 852 & English & Argument role type, entity type, co-referential text event \\
MultiHiEve \cite{ayyubi24bge} & Text + Video & News & 100,000 & English & Event-event relations \\
M-VAE \cite{ma25sms} & Text + Video & General & 1,680 & English & Natural language answer \\
MM Chinese EE \cite{zhang23mce} & Text + Speech & Broadcast news, newswire, weblog & 6,694 & Chinese & Event type \\
MUIE \cite{zhang24rea} & Text + Image + Audio & General & 3,000 & English & Event type, entity label, argument, role \\
\bottomrule
\end{tabular}
\label{tab:mmee-benchmarks}
\end{table*}

\textbf{ACE05}: The ACE 2005 dataset \cite{doddington04ace1} is the first large-scale, multilingual corpus to systematically define and annotate event tasks. It covers five main event types—Interaction, Movement, Transfer, Creation, and Destruction—each with detailed subtypes. Events are annotated with triggers, participants and their roles, and attributes such as time, location, instrument, and purpose, yielding a rich event structure. The dataset includes English, Chinese, and Arabic newswire, broadcast, and newspaper texts totaling about 500,000 words and thousands of event instances.

\textbf{Causal-TimeBank}: Causal-TimeBank \cite{mirza14acb} extends the TimeBank corpus with causality annotations across 183 news documents. It adds annotations for 137 events to the original 6{,}811, yielding 318 causal links and 171 causal signals. Compatible with TimeML temporal tags, it supports research on causality extraction and temporal–causal interaction analysis.

\textbf{HiEve}: The HiEve dataset \cite{glavavs14h} is a news event hierarchy corpus with 100 documents, 1,354 sentences, and 33,000 tokens, with an average of 32 events per document. It manually annotates spatiotemporal relations to form DAG-structured hierarchies and serves as the first public resource for event hierarchy modeling in extraction and summarization tasks.

\textbf{ERE}: The ERE dataset \cite{song15lre1}, developed by LDC under DARPA DEFT, is a multilingual corpus in English, Chinese, and Spanish. Light ERE provides simplified entity, relation, and event annotations, while Rich ERE adds Realis labels and Event Hoppers for event coreference. Covering newswire and forums, it supports large-scale knowledge-base construction and event extraction research.

\textbf{KBP2017}: The KBP 2017 dataset \cite{choubey17tak} includes both formal news and informal forum texts, challenging models to handle varied structures and noise. It supports tasks such as event trigger detection, type classification, realis identification, and document-level coreference, and serves as a key benchmark for cross-domain event extraction.

\textbf{MAVEN}: MAVEN \cite{wang20m} is a large, human-annotated event detection dataset from Wikipedia, with 4{,}480 documents, 118{,}732 event mentions, and 168 types—about twenty times larger than ACE 2005. With broad coverage and realistic long-tail, multi-event distributions, it is a key resource for general-domain event detection research.

\textbf{ECB+}: The ECB+ dataset \cite{cybulska14scn} extends the EventCorefBank (ECB) corpus with 502 additional news articles describing different instances of the same event types, increasing linguistic diversity and representativeness for studying cross-document event coreference in news texts.

For additional datasets and details, see Table~\ref{tab:ee-benchmarks}.

\subsection{Multimodal Datasets }

\textbf{imSitu}: The imSitu dataset was introduced in \cite{yatskar16srv} to support a structured understanding of what is happening in an image. It contains over 500 activities, 1,700 roles, 11,000 objects, 125,000 images, and 200,000 unique situations.
Each image is annotated with one or more ``situations'' defined by a verb and a set of semantic roles (e.g., Agent, Tool, Place) filled by noun entities grounded in the image. Notably, imSitu draws its role schema from linguistic resource FrameNet \cite{fillmore03bf}, and its noun entities from ImageNet \cite{russakovsky15ils}, enabling a rich yet consistent mapping between visual entities and semantic roles.

\textbf{SWiG}: 
The SWiG (Situations With Groundings) dataset \cite{pratt20gsr} extends the imSitu situation recognition benchmark by enriching it with explicit grounding annotations. While imSitu provides event categories and semantic role labels, SWiG further associates each argument with bounding-box annotations in the image, thereby linking semantic roles to concrete visual regions. In total, the dataset contains 126,102 images spanning 504 verbs, 278,336 bounding boxes, and more than 11,538 entity categories, covering a wide variety of actions and participants. This design enables models not only to recognize what event is happening and which entities are involved, but also to localize these entities in the visual scene, making it a comprehensive benchmark for studying structured visual event extraction and grounded situation recognition.

\textbf{VASR}: 
The {VASR} dataset \cite{bitton23vva} introduces a novel task of visual analogy of situation recognition. Using the imSitu situation recognition annotations, the authors automatically generated over 500,000 analogy instances of the form (A, A', B, B'), where the change from A to A' (e.g., agent, tool, or item role swapped) is mirrored by a corresponding change from B to B'. The silver-labeled analogies were further validated via crowdsourcing, yielding a gold-standard subset of 3,820 high-confidence analogies. 

\textbf{VidSitu}: The VidSitu dataset \cite{sadhu21vsr} is a large-scale benchmark for video situation recognition. It consists of about 29,000 ten-second clips sampled from around 3,000 movies, each densely annotated at 2-second temporal intervals. 
Every clip contains up to five salient events, and each event is labeled with a verb sense together with semantic roles grounded in video captions. 
The annotations also include entity co-reference across event segments within a video, as well as inter-event relations such as causation, contingency, and reaction. 
On average, each clip contains 4.2 unique verbs and 6.5 distinct entities, covering a verb vocabulary of roughly 1500 unique verbs (with over 200 verbs having at least 100 examples) and diverse noun entities
(5600 unique nouns with 350 nouns occurring in at least 100 videos). 

For additional datasets and details, see Table~\ref{tab:mmee-benchmarks}.

\subsection{Evaluation Metrics}
\label{sec:metrics}

We categorize evaluation methodologies into three primary domains: Event Extraction (EE), Event Relation Extraction (ERE), and Event Coreference Resolution (ECR).

\paragraph{Event Extraction Tasks}
Current literature typically evaluates EE performance under six distinct settings:

\begin{itemize}
    \item \textbf{Trigger Identification (Tri-I)}: the predicted trigger span exactly matches the gold span.
    \item \textbf{Trigger Classification (Tri-C)}: both the span and the event type are correct.
    \item \textbf{Argument Identification (Arg-I)}: the predicted argument span exactly matches the gold span.
    \item \textbf{Argument Classification (Arg-C)}: both the argument span and its role are correct.
    \item \textbf{Head Classification (Head-C)}: evaluates the correctness of predicted headwords (primarily for \textsc{Wikievent}).
    \item \textbf{Coref Match F1}: credits predictions where the argument belongs to the same coreference cluster as the gold entity.
\end{itemize}

To demonstrate the calculation protocol, we take \textbf{Trigger Identification (Tri-I)} as a representative example. Let $\mathcal{T}_{pred}$ denote the set of predicted trigger spans and $\mathcal{T}_{gold}$ denote the set of gold trigger spans. The Precision ($P$), Recall ($R$), and F1-score ($F_1$) are calculated as:

\begin{equation}
\begin{aligned}
P_{\text{Tri-I}} &= \frac{|\mathcal{T}_{pred} \cap \mathcal{T}_{gold}|}{|\mathcal{T}_{pred}|}, \\
R_{\text{Tri-I}} &= \frac{|\mathcal{T}_{pred} \cap \mathcal{T}_{gold}|}{|\mathcal{T}_{gold}|}, \\
F1_{\text{Tri-I}} &= \frac{2 \cdot P_{\text{Tri-I}} \cdot R_{\text{Tri-I}}}{P_{\text{Tri-I}} + R_{\text{Tri-I}}}.
\end{aligned}
\label{eq:tri_metrics}
\end{equation}

\noindent Metrics for the other five tasks (Tri-C, Arg-I, etc.) follow the same formulation, differing only in the matching criteria used to define the intersection set (True Positives).

\paragraph{Relation Extraction}
For Temporal, Causal, and Subevent relation extraction, the field adopts the standard micro-averaged $P$, $R$, and $F_1$ scores. The calculation mirrors Eq.~\ref{eq:tri_metrics}, where a True Positive is defined by the specific relation matching rules of the target dataset.

\paragraph{Event Coreference Resolution}
Event coreference is standardly evaluated using four cluster-based metrics:
\begin{itemize}
    \item \textbf{MUC} \cite{vilain95mtc}: A link-based metric measuring the minimum number of missing or spurious links.
    \item \textbf{B$^3$} \cite{bagga98ecd}: A mention-based metric averaging precision and recall over individual mentions.
    \item \textbf{CEAF$_e$} \cite{luo05crp}: An entity-based metric solving for the optimal one-to-one alignment between gold and system clusters.
    \item \textbf{BLANC} \cite{recasens11bir}: A metric designed to balance performance between coreference and non-coreference links.
\end{itemize}

\paragraph{Semantic and Generative Evaluation}
To address the limitations of Exact Span Matching, particularly for Generative Event Extraction (GEE), recent surveys and benchmarks have introduced semantic evaluation protocols. 
Common approaches include word-overlap metrics (e.g., BLEU, ROUGE), embedding distance (e.g., BERTScore), and LLM-based evaluation agents.

Notably, Lu et al. \cite{lu24bem} proposed \textsc{RAEE}, employing LLMs to compute semantic-level $F_1$ via adaptive prompting. 
Similarly, Lu et al. \cite{lu25s} introduced \textsc{SEOE}, a framework integrating multi-source ontologies to enable semantic matching across open-domain events. 
For argument evaluation, Fane et al. \cite{fane25bmb} developed \textsc{BEMEAE}, which combines deterministic text normalization with semantic similarity to better correlate with human judgments.
\begin{highlightbox}{Evaluation Gap: Exact Match vs. Semantic Utility}
\begin{compactlist}
    \item Exact Match ($F_1$) penalizes valid paraphrases, failing to reflect true model capability in generation.
    \item Semantic Evaluation shifts focus to utility and meaning equivalence, which is essential for assessing LLM outputs.
\end{compactlist}
\end{highlightbox}

\begin{table}[t]
\small
\rowcolors{2}{gray!10}{white}
\caption{Categorization of Tools}
\label{tab:tool}
\centering
\begin{tabular}{p{0.4\linewidth}p{0.5\linewidth}}
\toprule
\textbf{Tool} & \textbf{Category} \\
\midrule
DyGIE++~\cite{wadden19ere} & EE Toolkit \\
SpERT~\cite{eberts19sje} & EE Toolkit \\
OneIE~\cite{lin20jnm} & EE Toolkit \\
CasEE~\cite{sheng21cjl} & EE Toolkit \\
RESIN~\cite{wen21rds} & EE Toolkit \\
UIE~\cite{lu22usg} & EE Toolkit \\
InstructUIE~\cite{gui24ibi} & EE Toolkit \\
DeepKE~\cite{zhang22ddl} & EE Toolkit \\
OneEE~\cite{cao22oos} & EE Toolkit \\
DocEE~\cite{tong22dls} & EE Toolkit \\
SPEED++~\cite{parekh24sme} & EE Toolkit \\
OmniEvent~\cite{peng23ocf} & EE Toolkit \\
TextEE~\cite{huang23tbr} & EE Toolkit \& Evaluation scripts\\
ACE~\cite{doddington04ace} & Evaluation scripts\\
TAC-KBP~\cite{ji11kbp,surdeanu14oes} & Evaluation scripts\\
ERE~\cite{song15lre} & Evaluation scripts \\
MAVEN-ERE~\cite{wang22meu} & Evaluation scripts \\
BRAT~\cite{stenetorp12bwt} & Annotation Platform \\
RESIN~\cite{wen21rds}~\cite{stenetorp12bwt} & Annotation Platform \\
CollabKG~\cite{wei24clh} & Annotation Platform \\
LFDe~\cite{kan24llf} & Annotation Platform \\
\bottomrule
\end{tabular}
\end{table}

\subsection{Tools}
Over the past few years, a variety of reusable toolkits have been released to support event extraction, reflecting the steady maturation of research infrastructure. Early frameworks such as \textbf{DyGIE++}~\cite{wadden19ere} and \textbf{SpERT}~\cite{eberts19sje} introduced span-based modeling for entities, relations, and events. Building on this, \textbf{OneIE}~\cite{lin20jnm} provided a unified document-level pipeline that integrates entities, relations, and events.
Subsequent toolkits focused on specialization and accessibility. \textbf{CasEE}~\cite{sheng21cjl} proposed a cascade decoding framework for overlapping events, while \textbf{RESIN}~\cite{wen21rds} released a dockerized schema-guided pipeline for cross-document event extraction. 
In parallel, the focus shifted toward generalization. \textbf{UIE}~\cite{lu22usg} and its extension \textbf{InstructUIE}~\cite{gui24ibi} framed extraction as a generative task, enabling zero- and few-shot adaptation, while \textbf{DeepKE}~\cite{zhang22ddl} offered an open-source toolkit with ready-to-use pipelines covering entities, relations, and events. 
More recently, new platforms have targeted broader coverage and scalability. \textbf{OneEE}~\cite{cao22oos} and \textbf{DocEE}~\cite{tong22dls} addressed one-stage fast EE and document-level EE respectively. \textbf{SPEED++}~\cite{parekh24sme} introduced a multilingual EE framework with cross-lingual evaluation, and \textbf{OmniEvent}~\cite{peng23ocf} provided a large-scale open-source toolkit that integrates datasets, models, and evaluation scripts. \textbf{TextEE}~\cite{huang23tbr} consolidated these developments by offering a benchmark suite with runnable code and standardized evaluation, marking the current stage of maturity in EE toolkit development.

In parallel with toolkit development, the community has also established standardized evaluation scripts to ensure reproducibility and comparability. Early benchmarks such as the \textbf{ACE}~\cite{doddington04ace} program introduced task definitions and official scorers for precision, recall, and F1. The subsequent \textbf{TAC-KBP}~\cite{ji11kbp,surdeanu14oes} evaluations extended this tradition to large-scale knowledge base population, providing official evaluation code for entity linking, slot filling, and event extraction. To support richer annotation, the \textbf{ERE}~\cite{song15lre} framework unified entity, relation, and event labeling with corresponding scoring scripts. More recent benchmarks have followed the same practice. \textbf{MAVEN-ERE}~\cite{wang22meu} released a large-scale dataset with an official evaluation pipeline, while \textbf{TextEE}~\cite{huang23tbr} consolidated prior resources and offered publicly available scoring scripts for standardized re-evaluation. These evaluation packages, often distributed through dataset repositories or task organizers, have become a de facto standard in event extraction, enabling fair comparison across models and accelerating methodological progress.

Beyond toolkits and evaluation scripts, researchers have also developed annotation platforms and schema converters to facilitate dataset creation and adaptation. Early annotation tools such as \textbf{BRAT}~\cite{stenetorp12bwt} provided web-based interfaces for labeling entities, relations, and events, and have become widely adopted in constructing new corpora. Subsequently, schema-aware infrastructures were introduced to address heterogeneous annotation guidelines: \textbf{RESIN}~\cite{wen21rds} released a dockerized schema-guided pipeline for cross-document and cross-lingual information extraction, enabling alignment of entities, relations, and events across datasets. More recently, human–machine collaboration has been explored to further streamline annotation: \textbf{CollabKG}~\cite{wei24clh} proposes a cooperative framework for knowledge graph and event annotation, while \textbf{LFDe}~\cite{kan24llf} presents a lighter and more data-efficient workflow. Together, these platforms and converters reduce annotation costs and enhance dataset interoperability, providing crucial support for large-scale event extraction research. For clarity, Table~\ref{tab:tool} presents a categorical summary of existing event extraction resources.

\begin{figure*}[!t]
\centering
\includegraphics[width=1\textwidth]{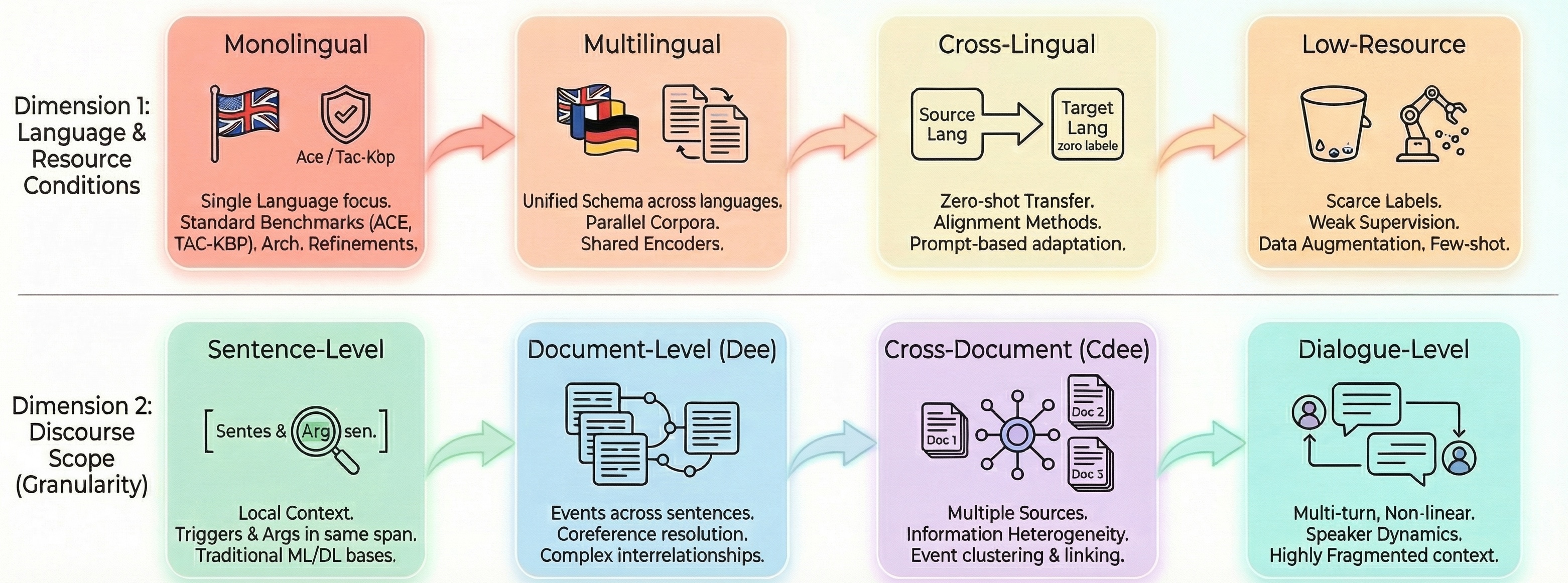}
\caption{
An overview of diverse research settings in Event Extraction, categorized into two primary dimensions: language and resource conditions, and discourse scope granularity.
} 
\label{fig:setting}
\end{figure*}

\section{EE under Diverse Settings}

\subsection{Language and Resource Conditions} 
In event extraction research, language setting defines the scope of applicability and generalization across linguistic contexts. This section categorizes EE research along four major dimensions: monolingual, multilingual, cross-lingual, and low-resource.

\paragraph{Monolingual}
Early EE research largely targets single-language settings—most often English—under standardized schemas and evaluation protocols such as ACE~\cite{doddington04ace}, TAC-KBP~\cite{ji11kbp,surdeanu14oes}, and ERE~\cite{song15lre}. Within this paradigm, modeling advances span formulation changes, decoders, and evaluation practice. Casting EE~\cite{liu20eem} as machine reading comprehension emphasizes span-centric reasoning and question-style conditioning to couple trigger decisions. Architectural refinements~\cite{sheng21cjl,zheng21ree} for overlapping or interdependent structures further improve extraction fidelity in single-language news and web domains. Beyond triggers, several works~\cite{xu24aac,liu23edl,he23rea} push document-level argument modeling with specialized structures or pipelines, reflecting a shift from sentence-local to discourse-level inference. New monolingual resources also expand the modality and language coverage while retaining a single-language assumption, e.g., speech-based EE benchmarks and non-English single-language corpora~\cite{wang24snb,nguyen24bpe}, which test robustness to acoustic noise and domain shift. As large language models (LLMs) become stronger few-shot reasoners, recent analyses~\cite{wei25hlp} probe preference biases and instruction sensitivity in argument extraction, highlighting evaluation nuances specific to monolingual pipelines.

\paragraph{Multilingual}
Multilingual EE aims to extract events across many languages with a unified schema and modeling backbone, often coordinating parallel or comparable corpora to stabilize cross-language label semantics. Resource design has therefore focused on schema alignment and language coverage. New multilingual datasets~\cite{veyseh22mnm} formalize cross-language consistency for triggers and provide broader typological diversity. On the modeling side, unified or massively multilingual encoders~\cite{huang22mgl,parekh24sme,jenkins23mml} serve as the backbone for schema-conditioned extraction, enabling a single model to operate across diverse languages while amortizing supervision. Beyond generic benchmarks, specialized multilingual corpora~\cite{borenstein23mee,menini24sfe}, such as historical news and domain-specific frames, stress-test schema transfer and semantic normalization when language-specific lexicalizations diverge.

\paragraph{Cross-lingual}
Cross-lingual EE trains on a source language, typically English and transfers to target languages with limited or no labels, emphasizing zero-shot transfer, alignment, and adaptation. Early work~\cite{lyu21zse} reframed EE through transfer-friendly interfaces to leverage cross-lingual sentence representations and reduce label mismatch. Transfer-based methods~\cite{vannguyen21ctl,elallaly21mmt,zhang22tls,zhou22mft,fincke22lmp}, including multi-task pretraining, priming, and the use of SRL as an auxiliary task, improve argument-role generalization by leveraging shared semantics. Prompt–based approaches~\cite{zhang22ezs,ma23ssp} paired with instruction-like conditioning further improve zero-shot robustness by decoupling surface forms from schema roles, while contextualized prompting addresses label drift across languages. Recent studies~\cite{cao23zsc,yi25ezs,cai23ezs} explicitly evaluate zero-shot cross-lingual structure recommendation and role transfer quality, clarifying where alignment succeeds or fails and how language-independent signals can mitigate lexical gaps. 

\paragraph{Low-resource}
When labeled data are scarce, EE performance hinges on how effectively models can construct surrogate supervision and generalize from a few examples. Data-centric methods~\cite{wang24tal,jin24sda} dominate, while targeted or schema-aware augmentation synthesizes diverse trigger contexts, improving coverage of rare types and roles. Few-shot document-level argument extraction~\cite{yang22fsd} explores how to transfer role semantics and coreference cues with minimal labels, often relying on discourse structure and prior knowledge to compensate for annotation sparsity. Orthogonally, lightweight or data-efficient pretraining~\cite{kan24llf} reduces reliance on large labeled corpora while preserving competitive accuracy, indicating that pretraining objectives and architecture choices can be tailored to low-resource regimes. Broader positioning work~\cite{kan24eee} emphasizes liberating EE from rigid data constraints by integrating weak supervision, augmentation, and careful evaluation design to better capture the realities of under-resourced languages and domains.

\subsection{Discourse Scope (Granularity)}
In event extraction (EE) research, the choice of discourse scope granularity defines the contextual boundaries of event analysis and directly impacts task complexity and application scenarios. This section categorizes EE granularity into four core levels: sentence/span-level, document-level, cross-document, and dialogue/conversation-level.

\paragraph{Sentence/Span-level}
\label{Discourse_Scope_sentence}
Event extraction focuses on the single sentence or phrase level, where event triggers and arguments often reside within the same sentence, making this task relatively simple. Early work utilized traditional machine learning methods to extract event triggers and classify arguments \cite{chen09lsi,li13jee,li14mle}. Other researchers also improved performance by incorporating lexical, syntactic, and semantic information into models \cite{hong11cei,mcclosky11eed,riedel11frj}. Subsequently, models such as CNNs and RNNs were introduced, enhancing the performance of event trigger identification and argument classification by enhancing their ability to model context \cite{chen15eed,li19bee,sha18jee}. At this time, researchers not only improved performance by incorporating syntactic information \cite{nguyen16jee,huang16lee,ece16zhao,liu18jme} but also utilized unsupervised (or weakly supervised) approaches to reduce reliance on large amounts of high-quality, manually annotated data \cite{ritter15wse, zhou15ufe,chen17ald}. With technological advancements, deep learning has gradually become mainstream. Researchers are generating data using pre-trained language models, reducing reliance on manually annotated data while also providing high-quality data \cite{yang19ept,chen24llm}. Furthermore, researchers have viewed EE as a natural language generation task \cite{lu21tcs,liu20eem}, for example, by asking questions to allow pre-trained language models to directly generate answers \cite{du20eea,li20eem,wei25hlp}, or by constructing specific prompt templates to achieve event extraction \cite{hsu22dde,peng24pem,lin25gge}.

\paragraph{Document-level}
Document-level event extraction (DEE) aims to extract structured event information from an entire document. Unlike sentence-level event extraction (SEE), DEE requires the integration of event information scattered across sentences in a document, which involves complex scenarios such as the scattered distribution of event arguments and interrelationships among multiple events. Early researchers used the same methods as SEE, such as traditional machine learning methods \cite{liao10dlc}, or incorporating syntactic information to enhance event extraction performance \cite{chen12jmc,yang16jee} and using supervised learning to automatically generate large amounts of annotated data \cite{yang18ddl}. However, this method cannot effectively handle complex issues such as the dispersion of arguments across sentences and the coexistence of multiple events. Generative methods \cite{yang21dle,sun24uum} are favored by researchers, who extract events by constructing and filling in specific prompt templates \cite{li21dle,ma22pep,liu22dpt}. With the rapid development of large language models, generative methods have received more attention in DEE \cite{zhang24uul,zhou24llt,xu25miu}.
\begin{highlightbox}{Scope: Local vs. Global Reasoning}
\begin{compactlist}
    \item Sentence-level extraction focuses on local syntax and identifying immediate triggers.
    \item Document-level extraction demands memory mechanisms to track entities and logic to resolve cross-sentence conflicts.
\end{compactlist}
\end{highlightbox}

\paragraph{Cross-document}
Cross-document event extraction (CDEE) is a method that integrates information from multiple documents to extract structured events. Unlike traditional document-level and sentence-level methods, CDEE aims to mine descriptions of the same event from different sources through collaborative analysis of multiple documents, ultimately forming a comprehensive and consistent view of the event. Compared to DEE and SEE, CDEE faces challenges such as information heterogeneity and inconsistency, as well as cross-document coreference resolution. Most existing research clusters documents with similar content through clustering or semantic similarity, then analyzes each cluster \cite{ji08ree,tsolmon14eem,intxaurrondo15dre,li16cge,jin17nfe}. In addition, some researchers have attempted to generate events directly from multiple documents using generative models \cite{liu19ode,wen21rds}, but generative approaches face higher time and space overhead when processing large numbers of documents simultaneously.

\paragraph{Dialogue/Conversation-level}
Conversation-level event extraction is the task of identifying and extracting event information from multi-turn conversations. 
While similar to DEE, which requires processing multiple sentences, conversation-level data differs in that it is nonlinear, colloquial, involves dynamic interactions between multiple actors, and is highly fragmented.
Event information is scattered across multiple conversation turns and depends on speaker identity and contextual intent. It also requires handling ellipsis, references, and conversational behavior variations. Currently, research on conversation-level event extraction is still in its infancy. Eisenberg et al. \cite{eisenberg20aep} constructed the Personal Events in Dialogue Corpus (PEDC) and proposed a method for automatically extracting personal events from conversations. They used a support vector machine (SVM) model for event classification and explored four feature types and different machine learning protocols. Sharif et al. \cite{sharif24eis} also constructed a conversation-level dataset, DiscourseEE, and proposed a relaxed matching evaluation method based on semantic similarity, addressing the problem of ``exact matching'' evaluation leading to severe underestimation of performance in generative event extraction.

\subsection{Vertical Domains}
Event extraction (EE) has been extensively studied in open domains, yet its deployment in vertical domains has become increasingly essential as real-world applications often operate within specialized contexts. Each domain is characterized by its own linguistic patterns, terminologies, and knowledge structures, leading to substantial variation in data distributions. As a result, models trained on generic corpora often face severe performance degradation when transferred to domain-specific texts, highlighting the need for domain adaptation and knowledge-guided learning strategies. At the same time, the growing prevalence of multimodal data—spanning text, images, and videos—has expanded the scope of EE beyond language alone, prompting researchers to explore how domain-specific and cross-modal cues can jointly enhance event understanding.

In this survey, we examine both general domain and vertical-domain event extraction. We include general domain EE not as a vertical domain in the strict sense, but as a complementary setting that deals with unrestricted event types and heterogeneous text sources, offering a valuable contrast to domain-constrained scenarios. Alongside general domain EE, we discuss representative vertical domains such as biomedical, financial, and legal EE, where rich domain knowledge and task-specific schemas drive distinctive modeling challenges and applications. Together, these perspectives provide a comprehensive view of how event extraction evolves from general-purpose frameworks toward specialized and multimodal intelligence.

\paragraph{General domain
}
General domain event extraction (EE) targets text without a pre-specified ontology, aiming to identify events across heterogeneous sources such as news articles, web pages, and social media streams. Unlike vertical domains, where schemas are tightly defined, general domain EE must cope with continuously emerging event types and evolving triggers. This diversity raises challenges in coverage, adaptability, and generalization. To address these issues, several benchmarks have been proposed. The ACE 2005 corpus \cite{doddington04ace} and its extensions through TAC KBP evaluations \cite{mitamura15otk} provided early large-scale resources. More recently, the MAVEN dataset \cite{wang20m} significantly extended coverage by annotating 168 event types across 118,000 documents, establishing the largest human-annotated general domain benchmark. The RAMS corpus \cite{ebner20msa} emphasized document-level arguments, highlighting the importance of contextual reasoning.

More recently, data augmentation with generated images and captions has been employed to enhance multimedia EE \cite{Du2023TrainingME}.
At the same time, Theia \cite{Moghimifar2023TheiaWS} proposed weakly supervised multimodal EE with incomplete annotations, while TSEE \cite{li23tsm} introduced a three-stream contrastive framework for text–video EE. On the language modeling side, instruction-tuned large language models \cite{Srivastava2025InstructionTuningLF} have been shown to significantly improve the flexibility of open-domain EE, and GLEN \cite{li23glen} extended the coverage of event types to thousands, pushing the boundary of schema scalability.

In addition to mainstream corpora, some studies have investigated unusual sources such as historical newspapers \cite{borenstein23mee} or video transcripts \cite{veyseh22eev}, highlighting the adaptability of open-domain EE to diverse modalities and data conditions. Collectively, these advances demonstrate that open-domain EE is moving toward schema-flexible, multimodal, and generalizable approaches, which are essential for real-world applications such as news monitoring, crisis management, and social media event tracking.

\paragraph{Biomedical/Clinical}

Biomedical and clinical event extraction (EE) aims to identify complex interactions among biomedical entities (e.g., genes, proteins, drugs, diseases) or clinically relevant events such as diagnoses, treatments, and adverse drug reactions. Compared with open-domain EE, this domain is characterized by lengthy terminology, frequent abbreviations, and nested entity structures, which together pose significant modeling challenges. Importantly, biomedical and clinical EE has received more attention and produced more work than many other domains. This is largely due to the abundance of publicly available textual resources such as PubMed abstracts, PMC full-text articles, and electronic health records, as well as pressing real-world needs in drug discovery, disease understanding, and healthcare. Furthermore, long-running benchmark campaigns such as the BioNLP Shared Tasks \cite{kim09obs} \cite{kim11o} \cite{bossy13bst} \cite{dele16obb} have consistently provided datasets and schemas that stimulate research in this area.

Among recent methods, DeepEventMine \cite{trieu20dee} introduced an end-to-end framework for nested biomedical events and remains a strong baseline.
Reinforcement learning has also been explored in biomedical event extraction. Zhao\cite{Zhao2020ANM} formulated multi-event extraction as a sequential decision process and incorporated external biomedical knowledge bases to guide argument detection. A subsequent study by Zhao\cite{Zhao2021AnIR} improved this RL framework through self-supervised data augmentation, enabling more robust learning under sparse annotation. These works illustrate how RL can be strengthened by domain knowledge and auxiliary supervision in biomedical settings.
Beyond reinforcement learning–based approaches, several recent studies have further diversified the methodological landscape of biomedical event extraction.
Fine-grained attention mechanisms have been explored to capture subtle semantic distinctions in complex biomedical interactions\cite{He2022ABE}. Tree-structured attentive models such as Child-Sum EATree-LSTMs \cite{Wang2023ChildSumEE} leverage syntactic hierarchies to enhance event representation. Knowledge-guided hierarchical graph networks \cite{Zhang2023BiomedicalEC} incorporate curated biomedical knowledge to improve causal relation extraction. Constraint-based multi-task frameworks like CMBEE \cite{Hu2024CMBEEAC} promote structural consistency by jointly learning triggers, arguments, and event schemas under explicit constraints. More recently, framing biomedical EE as a semantic segmentation problem \cite{Gao2024BiomedicalEE} has introduced a fully end-to-end and schema-flexible perspective, demonstrating strong performance in handling complex and low-resource scenarios.

\begin{figure*}[!t]
\centering
\includegraphics[width=1\textwidth]{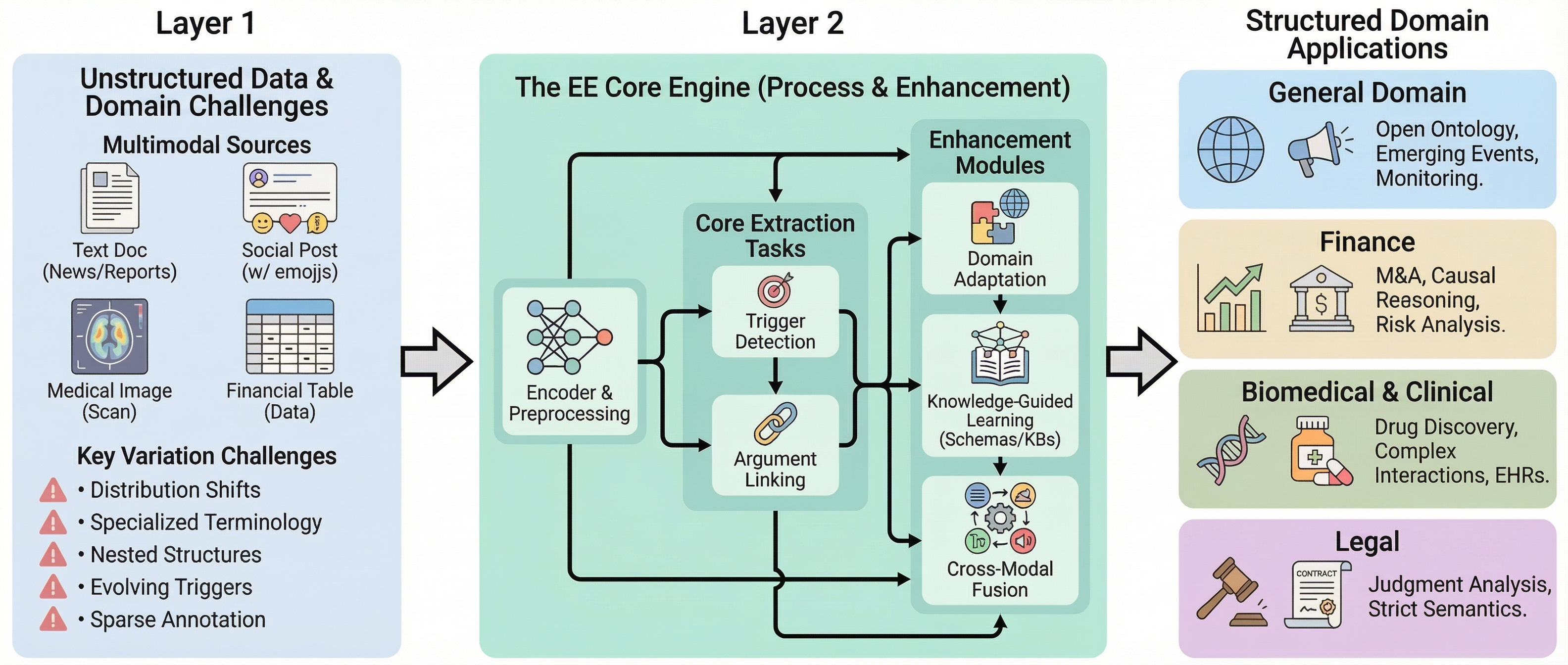}
\caption{
A three-layer architecture for Event Extraction applications in various domains. 
} 
\label{fig:domain}
\end{figure*}

Clinical event extraction has likewise attracted increasing attention. Datasets such as PHEE for pharmacovigilance \cite{sun22p}, discharge summaries annotated with adverse events \cite{Guellil2025AdverseEE}, and ACES for cohort event-stream data \cite{Xu2024ACESAC} provide valuable resources. Models like DICE \cite{ma23d} introduced data-efficient clinical EE with generative models, while disorder-aware attention mechanisms \cite{Yadav2020ExploringDA} improved recognition of clinical events from electronic health records. More recently, multilingual frameworks such as SPEED++ \cite{parekh24s} have expanded biomedical and epidemic event extraction to global contexts.

Collectively, these advances demonstrate that biomedical and clinical EE has evolved from feature-based methods on BioNLP benchmarks to diverse neural and generative approaches, supported by specialized datasets. The richness of resources, the continuity of community benchmarks, and the urgent societal needs explain why this domain has generated a larger body of work compared to others. It remains vital for applications such as biomedical knowledge base construction, drug discovery, pharmacovigilance, and clinical decision support.

\paragraph{Finance}
Financial event extraction (EE) focuses on identifying market- and economy-related events such as mergers and acquisitions, bankruptcies, stock fluctuations, policy changes, and equity pledges. Compared with open-domain settings, financial EE often requires reasoning over temporal and causal relations, since the sequence of events (e.g., policy issuance followed by market reaction) is essential for investment and risk analysis.

In terms of resources, financial EE exhibits a clear imbalance across languages. Chinese datasets are far more abundant, largely due to the availability of structured company announcements and financial disclosures in the Chinese market. Representative benchmarks include DCFEE \cite{yang18d}, the first large-scale document-level financial EE system based on distant supervision, and Doc2EDAG \cite{wang22d}, which modeled document-level financial events as directed acyclic graphs. Subsequent datasets such as FinEvent \cite{peng22ric}, CFinDEE \cite{zhang24ccf}, and CFERE \cite{wan23cmt} further enriched fine-grained and multi-type financial event coverage. More recently, OEE-CFC \cite{wan24o} explored open EE from Chinese financial commentary, and Probing into the Root \cite{Chen2021ProbingIT} provided a benchmark for reasoning about the causes of structural events.

By contrast, English financial EE datasets remain scarce. While efforts such as automatic extraction of financial events from news \cite{nguyen16jee} and the Event Causality in Finance benchmark \cite{Mariko2020TheFD} have begun to fill the gap, most English studies still rely on self-constructed corpora from financial news (e.g., Reuters, Bloomberg) that are not fully open due to licensing restrictions. As a result, many methodological innovations in financial EE—such as reinforcement learning, graph-based document modeling, or causal reasoning—have been validated primarily on Chinese datasets, with only limited evaluation on English data.

Overall, financial EE has grown from early document-level systems to datasets and models capable of fine-grained, causal, and open extraction. The uneven resource distribution between Chinese and English underscores both the maturity of research in the Chinese community and the need for broader multilingual benchmarks. These advances highlight the practical value of financial EE for applications such as investment decision-making, risk management, and financial knowledge graph construction.

\paragraph{Legal}
Legal event extraction seeks to extract structured events from legal texts such as court judgments, contracts, case reports, and other judicial documents. Compared with general open-domain or vertical domains, the legal domain imposes stricter semantics, domain-specific schemas, and high requirements for interpretability and correctness. The scarcity of publicly annotated legal event datasets has constrained model development, but recent work is beginning to fill this gap.

One significant dataset is LEVEN \cite{yao22lls}, which provides 8,116 Chinese legal documents annotated with 150,977 event mentions across 108 event types, laying a foundation for scale in legal event detection. To further enrich granularity, LEEC \cite{xue23lle} defines an extensive label system of 159 elements for criminal documents, supporting fine-grained extraction tasks. On the methodological side, Hierarchical legal event extraction via Pedal Attention Mechanism \cite{shen20hcl} introduced hierarchical features and attention mechanisms specific to Chinese legal events. More recently, Event Grounded Criminal Court View Generation \cite{yue24egc} combines event extraction and legal document generation using LLMs, injecting fine-grained legal events into court view summarization.
Earlier work such as Event Extraction for Legal Case Building and Reasoning \cite{Lagos2010EventEF} laid the conceptual groundwork for event-based legal case analysis. These developments illustrate a trajectory: from entity/event detection toward integrated event-centered legal reasoning, generation, and retrieval. Applications include legal knowledge graph construction, case retrieval, judgment summarization, and intelligent legal assistants.

\section{Open Challenges and Future Directions}
\label{sec:future}

The advent of generative LLMs and MLLMs has not merely pushed the performance boundaries on standard EE benchmarks; it has fundamentally precipitated a paradigm shift in how we conceptualize the task itself. We argue that EE is undergoing a critical transition from a static, sentence-level information extraction problem to a dynamic, document-level knowledge acquisition process intended for intelligent systems. Moving beyond the traditional limitations of span-based accuracy within isolated sentences, we identify six transformative directions that represent the true frontier of this field, demanding deeper investigation and architectural innovation.

\subsection{Agentic Perception} 

The traditional view of event extraction as a pipeline to populate static knowledge bases is increasingly insufficient in the era of autonomous agents. An intelligent agent operating in a complex environment does not need a static snapshot of history; it requires a continuously updated understanding of state changes and evolving situations to inform its actions. In this context, EE must be reimagined as the primary ``perception module'' intended to digest continuous text streams (e.g., news feeds, operational logs, dialogue history) and convert them into structured observations. The fundamental challenge shifts from simply identifying that an event occurred to understanding its implications for the agent's current state.

Future research must focus on integrating extracted events directly into the agent's cognitive architecture, specifically its episodic memory and world model. This means moving away from generating isolated JSON objects toward maintaining a coherent, temporally linked narrative of evolving realities. A critical research direction is developing mechanisms where newly extracted events can trigger updates to previously stored information, resolve conflicting observations over time, and directly feed into downstream modules responsible for causal reasoning, planning, and decision-making under uncertainty. The goal is to transform text into actionable situational awareness.
\begin{highlightbox}{Future Vision: Static vs. Dynamic}
\begin{compactlist}
    \item Static extraction traditionally populates offline databases from historical text.
    \item Dynamic perception serves as an Agent’s Memory, continuously updating state from streaming data.
\end{compactlist}
\end{highlightbox}

\subsection{Neuro-Symbolic Reasoning}
While LLMs exhibit impressive ``System 1'' intuition in identifying potential triggers and arguments based on surface-level semantic correlations, they remain fundamentally prone to probabilistic failures. They often struggle with complex structural constraints (e.g., an entity cannot be both an 'Attacker' and a 'Victim' in the same 'Attack' event unless specified) and suffer from factual hallucinations, generating plausible-looking but incorrect structures. The core limitation is the lack of explicit reasoning mechanisms to verify that the generated structure adheres to logical and ontological rules inherent to the domain.

To achieve robust and trustworthy extraction, a critical frontier is instilling ``System 2'' capabilities through neuro-symbolic integration. This goes beyond simply prompting an LLM to ``think step-by-step.'' Future frameworks should explore utilizing event schemas and logical rules not just as data preprocessing steps, but as hard constraints during the decoding process (e.g., constrained beam search via automata) to physically prevent invalid structures. Furthermore, we envision architectures that incorporate an intrinsic ``critic'' module—potentially a separate, logic-driven system—that deliberately verifies the generated event arguments against source evidence before final output, trading inference speed for structural guarantee.
\begin{highlightbox}{Future Direction: System 1 vs. System 2}
\begin{compactlist}
    \item System 1 (LLMs) provides probabilistic intuition but struggles with complex structural constraints.
    \item Future System 2 frameworks integrate Neuro-Symbolic verification to enforce logical validity and reliability.
\end{compactlist}
\end{highlightbox}

\subsection{Interactive Open-World Discovery}
The reliance on rigid, pre-defined ontologies (such as ACE or ERE) severely curtails the real-world applicability of EE systems. The long tail of complex scenarios contains myriad event types that cannot be anticipated ab initio. While recent zero-shot and few-shot approaches show promise, they often still operate under a closed-world assumption where the new types are known to the user beforehand. The true challenge lies in handling unknown unknowns—enabling systems to proactively identify clusters of information that indicate previously undefined event types evolving in the data.

The future paradigm must shift from passive extraction based on fixed instructions to interactive knowledge discovery. Future systems should not merely output a best guess when faced with ambiguity; they should possess the meta-cognitive ability to recognize their uncertainty and engage in clarification dialogues with human users (e.g., Does this text describe a new type of `Cyberattack', or is it a sub-event of 'Fraud'?). This collaborative approach allows the model to incrementally build and refine its own schema repository, evolving from a static classifier into an adaptive, lifelong-learning knowledge acquisition engine.
\begin{highlightbox}{Future Paradigm: Passive vs. Interactive}
\begin{compactlist}
    \item Passive extraction operates on fixed ontologies, forcing ambiguous inputs into pre-defined slots.
    \item Interactive discovery possesses meta-cognition to recognize uncertainty and proactively query users to learn new types.
\end{compactlist}
\end{highlightbox}

\subsection{Cross-Document Synthesis}
A significant disconnect exists between academic benchmarks, which predominantly focus on sentence-level extraction, and real-world information dynamics. Complex events—such as a pandemic outbreak, a corporate merger, or a geopolitical conflict—are rarely encapsulated within a single paragraph. Their core arguments (who, when, where, why) are fragmented across long documents or scattered among disparate sources published over extended periods. Current LLM-based approaches, despite larger context windows, struggle to synthesize conflicting information and maintain coherence over long horizons due to the lost-in-the-middle phenomenon and attention dilution.

Addressing this requires architectural innovations geared toward massive-scale context integration. Future research needs to develop specialized Retrieval-Augmented Generation (RAG) systems optimized for structured event data, capable of retrieving not just relevant sentences, but related existing event structures to inform current extraction. Furthermore, cracking cross-document event coreference resolution and temporal ordering is paramount. A sophisticated system must be able to link an arrest mentioned in today's news with an investigation mentioned last week, recognizing them as parts of the same larger event chain, rather than treating them as isolated data points.

\subsection{Physically Grounded World Models}
Current multimodal event extraction efforts often suffer from a shallow alignment limitation, where visual data is treated merely as supplementary signals to disambiguate textual entities (e.g., bounding box matching). This misses the profound potential of multimodal data: grounding language in physical reality. Text is inherently lossy; it rarely explicitly states common-sense physical details (e.g., that a break event implies an instrument and an irreversible state change). Relying solely on text limits the depth of understanding causal dynamics.

The next generation of multimodal models must leverage vast amounts of video data to learn the intuitive physics and temporal cause-and-effect chains of the real world. By pre-training on visual dynamics, future models should be able to infer implicit arguments that are visually obvious but textually unstated. For instance, presented with text about a car accident, a physically grounded model should implicitly understand the likely involvement of high velocity and impact forces, allowing it to predict potential consequences (e.g., injury, damage) even if unmentioned. This bridging of linguistic semantics with physical scene understanding is key to deeper event comprehension.
\begin{highlightbox}{Multimodal Evolution: Alignment vs. Grounding}
\begin{compactlist}
    \item Shallow alignment uses visual data merely as auxiliary signals to locate or disambiguate textual entities.
    \item Physical grounding learns intuitive physics from video to infer implicit causes and consequences unstated in text.
\end{compactlist}
\end{highlightbox}

\subsection{Utility-Driven Evaluation}
As EE shifts toward generative paradigms, traditional exact-match metrics like F1 scores based on character offsets are becoming functionally obsolete, penalizing valid paraphrases and failing to capture structural correctness. While shifting toward LLM-based semantic evaluation shows promise, it introduces new challenges regarding evaluator bias and reproducibility that must be rigorously studied. The community urgently needs standardized, trustworthy protocols for assessing semantic equivalence in structured outputs.

More critically, evaluation must expand along two neglected dimensions: utility and trustworthiness. Firstly, intrinsic metrics should be complemented by extrinsic \textit{utility-based evaluation}, measuring the quality of extraction by its tangible impact on downstream applications (e.g., Does this EE system measurably improve the accuracy of a quantitative financial forecasting model?). Secondly, for real-world deployment, metrics for \textit{calibration and uncertainty} are essential. A useful system must not only extract facts but also reliably signal its own confidence levels, knowing when it does *not* know, which is vital for ensuring safety and building user trust in critical domains.

\section{Conclusion}
This survey traces the evolution of Event Extraction (EE) and outlines promising directions for future work. We begin by defining the core tasks in EE, covering a spectrum from traditional text-based extraction to complex multimodal scenarios, and explore their applications across various domains. The review then maps out the field's methodological timeline, from rule-based systems and classical machine learning to deep learning and the current LLM-driven paradigm. We also provide a detailed analysis of system architectures, feature enhancement techniques, and essential resources like datasets, metrics, and toolkits. In the current stage, the arrival of Large Language Models has fundamentally changed the field, opening up new possibilities while introducing significant challenges. Looking ahead, we identify critical research frontiers, including the extraction of implicit events, the improvement of cross-modal alignment, and the development of reliable generative models. We hope this comprehensive overview serves as a clear guide to the field's development, inspiring innovative work that pushes the boundaries of event understanding.

\newpage
\newpage

\bibliographystyle{IEEEtran}
\bibliography{ref}
\nocite{*}

\end{document}